\pdfoutput=1
\documentclass[runningheads]{llncs}
\usepackage{graphicx}

\usepackage{tikz}
\usepackage{comment}
\usepackage{amsmath,amssymb} 
\usepackage{color}
\usepackage{multirow}
\usepackage{tabularx}
\usepackage{booktabs}
\usepackage{graphicx}
\usepackage{floatrow}
\usepackage{caption}
\usepackage{hyperref}

\newfloatcommand{figurebox}{figure}[\nocapbeside][\dimexpr(\textwidth-\columnsep)/2\relax]
\newfloatcommand{tablebox}{table}[\nocapbeside][\dimexpr(\textwidth-\columnsep)/2\relax]

\usepackage[accsupp]{axessibility}  

\usepackage{algorithm}
\usepackage{algorithmic}
\usepackage{subfigure}

\begin{document}
\pagestyle{headings}
\mainmatter
\def\ECCVSubNumber{728}  

\title{CLOSE: Curriculum Learning On the Sharing Extent Towards Better One-shot NAS} 

\titlerunning{CLOSE}
%

\author{%
  Zixuan Zhou\inst{13}\thanks{Equal contribution.} \and Xuefei Ning\inst{12}$^{*}$ \and Yi Cai\inst{1} \and Jiashu Han\inst{1} \and \\ Yiping Deng\inst{2} \and Yuhan Dong\inst{3} \and Huazhong Yang\inst{1} \and Yu Wang\inst{1}$^{\dagger}$%
}
\authorrunning{Z. Zhou, X. Ning et al.}
%
\institute{Department of Electronic Engineering, Tsinghua University \email{$^*$zhouzx21@mails.tsinghua.edu.cn, $^*$foxdoraame@gmail.com, $^{\dagger}$yu-wang@tsinghua.edu.cn} \and Huawei TCS Lab \and Tsinghua Shenzhen International Graduate School}
\maketitle

\begin{abstract}
One-shot Neural Architecture Search (NAS) has been widely used to discover architectures due to its efficiency. However, previous studies reveal that one-shot performance estimations of architectures might not be well correlated with their performances in stand-alone training because of the excessive sharing of operation parameters (i.e., large sharing extent) between architectures. Thus, recent methods construct even more over-parameterized supernets to reduce the sharing extent. But these improved methods introduce a large number of extra parameters and thus cause an undesirable trade-off between the training costs and the ranking quality. To alleviate the above issues, we propose to apply Curriculum Learning On Sharing Extent (CLOSE) to train the supernet both efficiently and effectively. Specifically, we train the supernet with a large sharing extent (an easier curriculum) at the beginning and gradually decrease the sharing extent of the supernet (a harder curriculum). To support this training strategy, we design a novel supernet (CLOSENet) that decouples the parameters from operations to realize a flexible sharing scheme and adjustable sharing extent. 
Extensive experiments demonstrate that CLOSE can obtain a better ranking quality across different computational budget constraints than other one-shot supernets, and is able to discover superior architectures when combined with various search strategies. Code is available at \url{https://github.com/walkerning/aw_nas}. 
\end{abstract}

\section{Introduction}

\label{sec:intro}

Neural Architecture Search (NAS)~\cite{zoph2016neural} has achieved great success in automatically designing deep neural networks (DNN) in the past few years. However, traditional NAS methods are extremely time-consuming for discovering the optimal architectures, since each architecture sampled in the search process needs to be trained from scratch separately. To alleviate the severe problem of search inefficiency,
one-shot NAS proposes to share operation parameters among candidate architectures in a ``supernet'' and train this supernet to evaluate all sampled candidate architectures~\cite{brock2017smash,pham2018efficient,bender2018understanding,liu2018darts}, which reduces the overall search cost from thousands of GPU days to only a few GPU hours. 

Despite its efficiency, previous studies reveal that one-shot NAS suffers from the poor ranking correlation between one-shot estimations and stand-alone estimations, which leads to 
unfair comparisons between 
the candidate architectures~\cite{luo2019understanding,sciuto2019evaluating,zela2020bench,ning2020surgery}.  Ning et al.~\cite{ning2020surgery} give some insights on the failure of one-shot estimations. They conclude that one of the main causes of the poor ranking quality is the \textbf{large sharing extent of the supernet}. Several recent studies try to improve one-shot NAS by addressing the large sharing extent issue. Zhao et al.~\cite{zhao2021few} confirm the negative impact of co-adaption of parameters in one supernet. Thus, they split the whole search space into several smaller ones, and train a supernet for each subspace. Su et al.~\cite{su2021k} reveal that only one copy of parameters is hard to be maintained for massive architectures. Therefore, they duplicate each parameter of the supernet into several copies, train and then estimate with all the duplicates. However, these improved methods introduce a large number of parameters, which barricades the supernet training.  
As a result, they have to make a trade-off between the training costs and the ranking quality.

In this paper, we propose to adopt the Curriculum Learning On Sharing Extent (CLOSE) to train the one-shot supernet efficiently and effectively. 
The underlying intuition behind our method is that training with a large sharing extent can \textbf{efficiently bootstrap} the supernet, since the number of parameters to be optimized is much smaller. While in the later training stage, using a supernet with a smaller sharing extent (i.e., a more over-parameterized supernet) can \textbf{improve the saturating ranking quality}.
Thus, CLOSE uses a relatively large sharing extent in the early training stage of the supernet, then gradually decreases the supernet sharing extent.
To support this training strategy, we design a new supernet with an adjustable sharing extent, namely CLOSENet, of which the sharing extent can be flexibly adjusted in the training process. 
The difference between CLOSENet and the vanilla supernet is that, the construction of vanilla supernets presets the sharing scheme between any architecture pairs, i.e., designates which parameter is shared by which operations in different architectures. In contrast, CLOSENet could flexibly adjust the sharing scheme and extent between architecture pairs, during the training process.

In summary, the contributions of our work are as follows:
\begin{enumerate}
    \item We propose to apply Curriculum Learning On Sharing Extent (CLOSE) to efficiently and effectively train the one-shot supernet. Specifically, we use a larger sharing extent in the early stages to accelerate the training process, and gradually switch to smaller ones to boost the saturating performances.
    \\
    \item To fit the CLOSE strategy, we design a novel supernet (CLOSENet) with an adjustable sharing extent. Different from the vanilla supernet with an unadjustable sharing scheme and sharing extent, CLOSENet can flexibly adapt its sharing scheme and sharing extent during the training process.
    \\
    \item Extensive experiments on four NAS benchmarks show that CLOSE can achieve a better ranking quality under any computational budgets. When searching for the optimal architectures, CLOSE enables one-shot NAS to find superior architectures compared to existing one-shot NAS methods.
    
\end{enumerate}

\section{Related Work}

\subsection{One-shot Neural Architecture Search (NAS)}
Neural Architecture Search (NAS) is proposed to 
find optimal architectures automatically. However, the vanilla sample-based NAS methods~\cite{zoph2016neural} are extremely time-consuming. 
To make it more efficient, Pham et al.~\cite{pham2018efficient} propose the parameter sharing technique by constructing an over-parameterized network, namely supernet, to share parameters among the candidate architectures. 
Based on the parameter sharing technique, various one-shot NAS methods are proposed to efficiently search for optimal architectures by only training ``one'' supernet. Bender et al.~\cite{bender2018understanding} propose to directly train the whole supernet with a path-dropout strategy. 
Liu et al.~\cite{liu2018darts} develop a differentiable search strategy and use it in conjunction with the parameter sharing technique. 
Guo et al.~\cite{guo2020single} propose to separate the stages of supernet training and architecture search. 

\subsection{Weakness and Improvement of One-shot NAS}
\label{sec:improvement}
Despite its high efficiency, one-shot NAS suffers from the poor ranking correlation between the architecture performances using one-shot training and stand-alone training.
Sciuto et al.~\cite{sciuto2019evaluating} discover that the parameter-sharing rankings do not correlate with stand-alone rankings by conducting a series of experiments in a toy search space.
Zela et al.~\cite{zela2020bench} confirm the poor ranking quality in a much larger NAS-Bench-1shot1 search space. Luo et al.~\cite{luo2019understanding} make a further investigation of the one-shot NAS, and attribute the poor ranking quality to the insufficient and imbalanced training, and the coupling of training and search phases. Ning et al.~\cite{ning2020surgery} provide comprehensive evaluations on multiple NAS benchmarks, and conclude three perspectives to improve the ranking quality of the one-shot NAS, i.e., reducing the temporal variance, sampling bias or parameter sharing extent. 

Recent studies adopt the direction of sharing extent reduction to improve the one-shot NAS. Ning et al.~\cite{ning2020surgery} prune the search space to reduce the number of candidate architectures, and reveal the improvement of the ranking quality in the pruned search space. 
But the ranking quality in the overall search space is not improved.
Zhao et al.~\cite{zhao2021few} propose \textit{Few-shot NAS} to split the whole search space into several subspaces, and train a single supernet for each subspace. Su et al.~\cite{su2021k} propose \textit{K-shot NAS} to duplicate each parameter of the supernet into several copies, and estimate architectures' performances with all of them. However, these two methods reduce sharing extent with even more over-parameterized supernets, which brings extra computational costs. 

\subsection{Curriculum Learning}
\label{sec:rw_cl}
Bengio et al.~\cite{bengio2009curriculum} first propose curriculum learning (CL) strategy based on the learning process of humans and animals in the real world. The basic idea of the CL strategy is to guide models to learn from easier data (tasks) to harder data (tasks). In the past few years, many studies have successfully applied CL strategy in various applications~\cite{guo2018curriculumnet,jiang2014easy,platanios2019competence,tay2019simple,ren2018self,gong2019multi,guo2020breaking}, and demonstrated that CL can improve the models' generalization capacity and convergence speed. 
Besides common CL methods that adjust the data, there exist CL methods that conduct curriculum learning on the model capacity. Karras et al.~\cite{karras2017progressive} propose to progressively increase the model capacity of the GAN to speed up and stabilize the training. Soviany et al.~\cite{soviany2022curriculum} propose a general CL framework at the model level that adjusts the curriculum by gradually increasing the model capacity.

\subsection{NAS Benchmarks}

NAS benchmarks enable researchers to reproduce the NAS experiments easily and compare different NAS methods fairly. 
NAS-Bench-201~\cite{dong2020bench} constructs a cell-based NAS search space containing 15625 architectures and provides 
their complete training information. 
NAS-Bench-301~\cite{siems2020bench} uses a surrogate model to predict the performances of approximately $10^{18}$ architectures in a more generic search space, with the stand-alone performance of 60k landmark architectures.
Different from NAS-Bench-201 and NAS-Bench-301 that focus on topological search spaces, 
NDS~\cite{radosavovic2020designing} provides benchmarks on two non-topological search spaces (e.g., ResNet~\cite{he2016deep} and ResNeXt~\cite{xie2017aggregated}). 

\section{Method}
\label{sec:method}

\subsection{Motivation and Preliminary Experiments}
\label{subsec:motivation}
In one-shot NAS, many operations in different architectures share the same parameter, while their desired parameters are not necessarily the same. The excessive sharing of parameters, i.e., the large sharing extent, has been widely regarded as the most important factor causing the unsatisfying performance estimation~\cite{benyahia2019overcoming,zhang2020overcoming,ning2020surgery,zhao2021few,su2021k}.
The most recent studies~\cite{zhao2021few,su2021k} improve the ranking quality by reducing the sharing extent. But their methods cause an inevitable trade-off between the training cost and ranking quality at the same time.

\begin{figure}[htb]
  \begin{center}
    \subfigure{
      \includegraphics[width=0.39\linewidth]{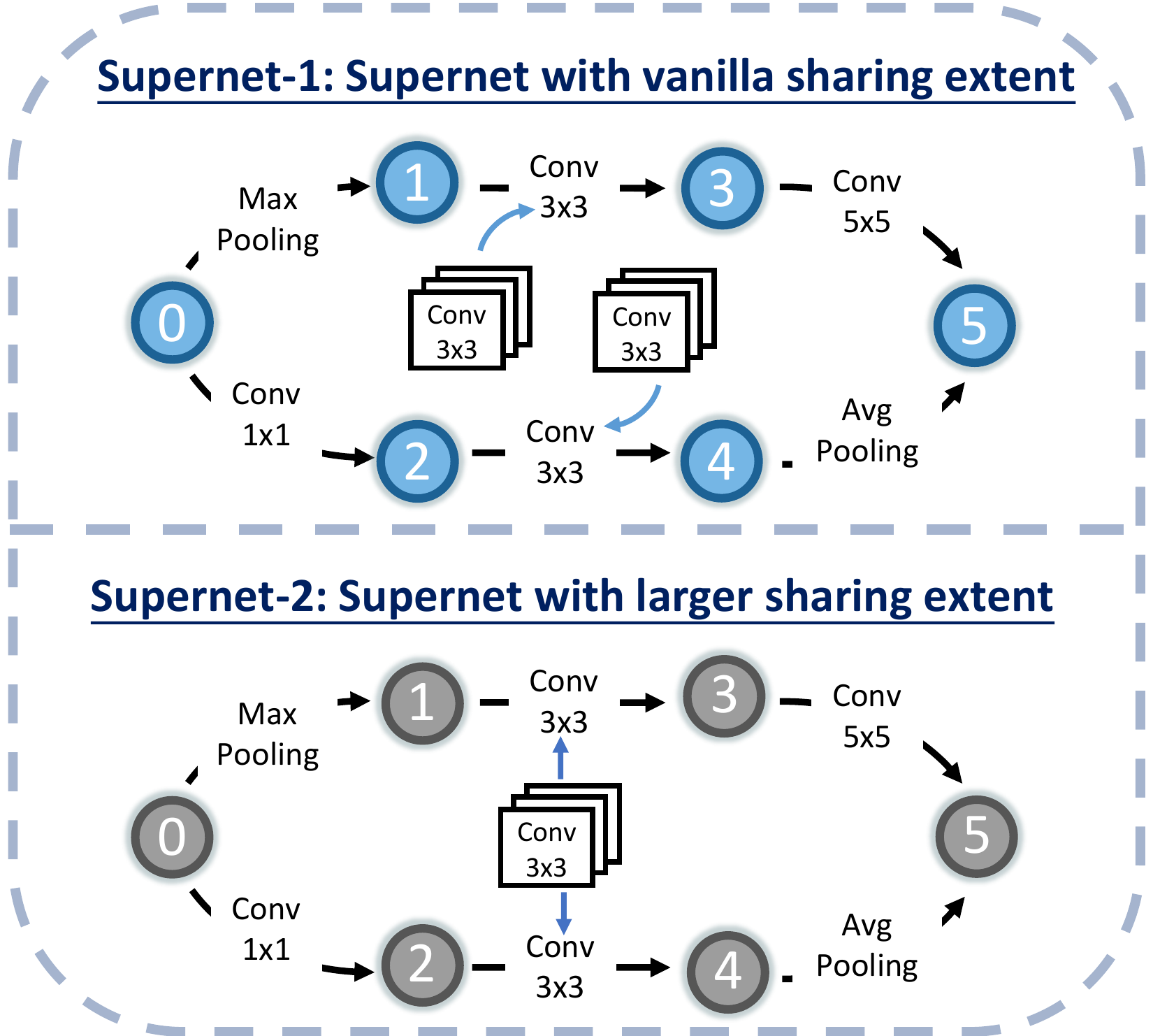}}
    \subfigure{
      \includegraphics[width=0.58\linewidth]{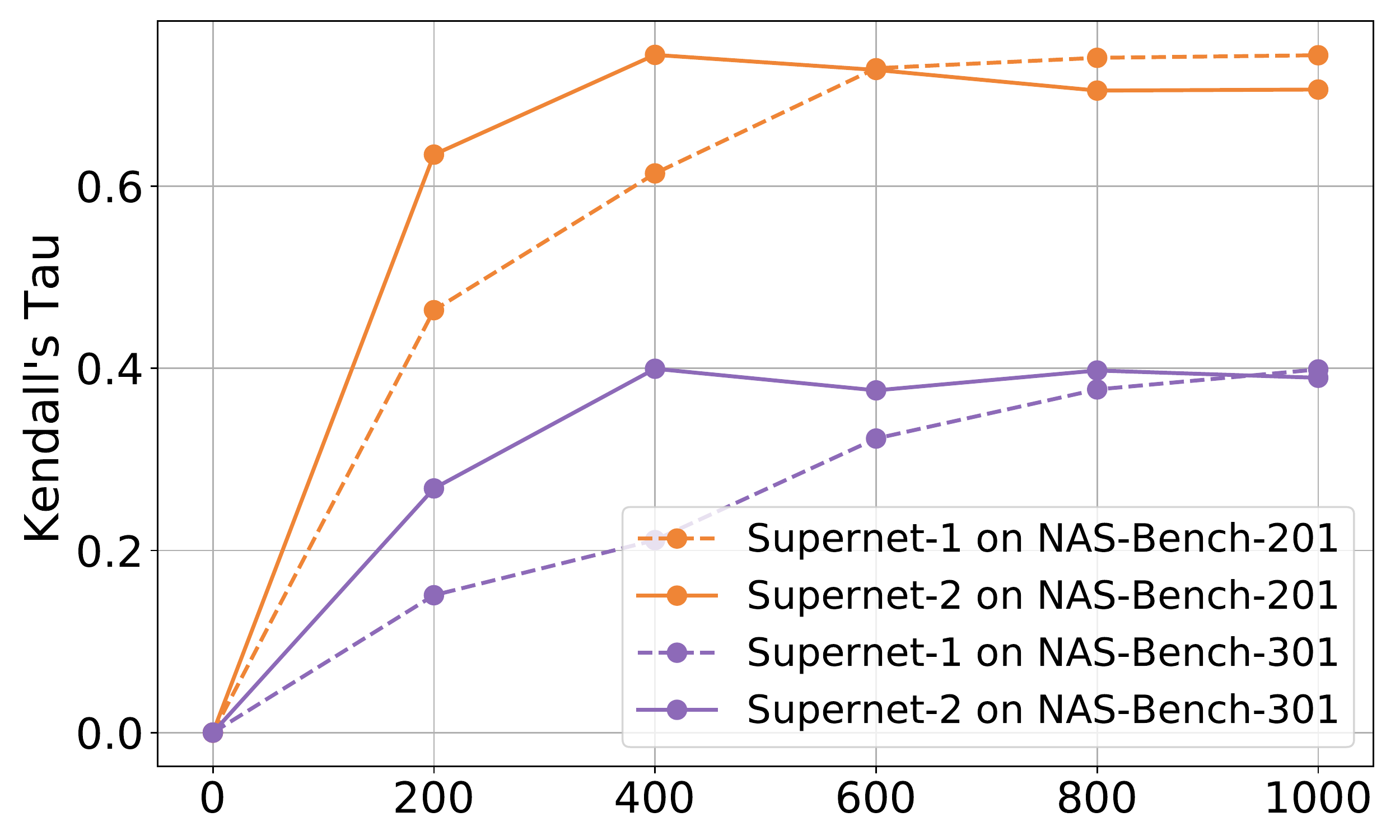}}
    \caption{Comparison of supernets with small and large sharing extent. The left part shows a cell-architecture that uses Supernet-1 (top) and Supernet-2 (bottom). The right part shows the Kendall's Tau of Supernet-1 and Supernet-2 throughout the training process on NAS-Bench-201 and NAS-Bench-301.}
    \label{fig:close1000epoch}
  \end{center}
\end{figure}

Supernets with larger sharing extents (i.e., more parameters) are easier to train in the early training stage. We verify this statement with an experiment on two popular cell-based NAS benchmarks. 
We construct two supernets with different sharing extents: Supernet-1 (a cell shown in top-left of Fig.~\ref{fig:close1000epoch}) is a vanilla supernet adopted by many one-shot NAS methods (e.g., DARTS~\cite{liu2018darts}), in which the compound edges in one cell use different copies of parameters. While Supernet-2 (a cell shown in the bottom-left of Fig.~\ref{fig:close1000epoch}) shares only one copy of parameters for all the compound edges in each cell, which leads to a much larger sharing extent than Supernet-1. 
For example, the parameters of $conv_{3\times 3}$ in edge (1,3) and (2,4) are different when using Supernet-1, but the same when using Supernet-2. We train them to convergence and use Kendall's Tau (see Sec.~\ref{sec:evalranking} for definition), to evaluate the ranking correlation between the estimated performances by the supernets and the groud-truth performances.

Fig.~\ref{fig:close1000epoch} (right) shows that, on one hand, using a smaller sharing extent (more parameters, larger supernet capacity) can alleviate the undesired coadaptation between architectures, and has the potential to achieve higher saturating performances. The Kendall's Tau of Supernet-2 is slightly worse than that of Supernet-1 when the supernets are relatively well-trained (epoch 800 to 1000). 
On the other hand, training the supernet with higher sharing extent than the vanilla one (fewer parameters, smaller supernet capacity) greatly accelerates the training process of parameters, and help the supernet obtain a good ranking quality faster. In the early stage of the training process (epoch 0 to 600), Supernet-2 has a much higher Kendall's Tau than Supernet-1. 

Based on the above results and analysis, \textbf{a natural idea to achieve a win-win scenario of supernet training efficiency and high ranking quality is to adapt the sharing extent during the supernet training process}. 
We can draw parallels between this idea and the CL methods on model capacity~\cite{karras2017progressive,soviany2022curriculum} (see Sec.~\ref{sec:rw_cl}), as they all progressively increase the capacity of the model or supernet to achieve training speedup and better performances in the mean time. 
Based on the above idea, we propose to employ Curriculum Learning On the Sharing Extent (CLOSE) of the supernet. And to enable the adaption of the sharing extent during the training process,
we design a novel supernet, CLOSENet, whose sharing extent can be easily adjusted. 

In the following, we first demonstrate the construction of CLOSENet in Sec.~\ref{subsec:closenet}. Then, in Sec.~\ref{subsec:close}, we describe CLOSE with some necessary training techniques to achieve the best ranking quality.

\subsection{CLOSENet: A Supernet with An Adjustable Sharing Extent}
\label{subsec:closenet}

\begin{figure}[htb]
  \begin{center}
    \includegraphics[width=0.90\linewidth]{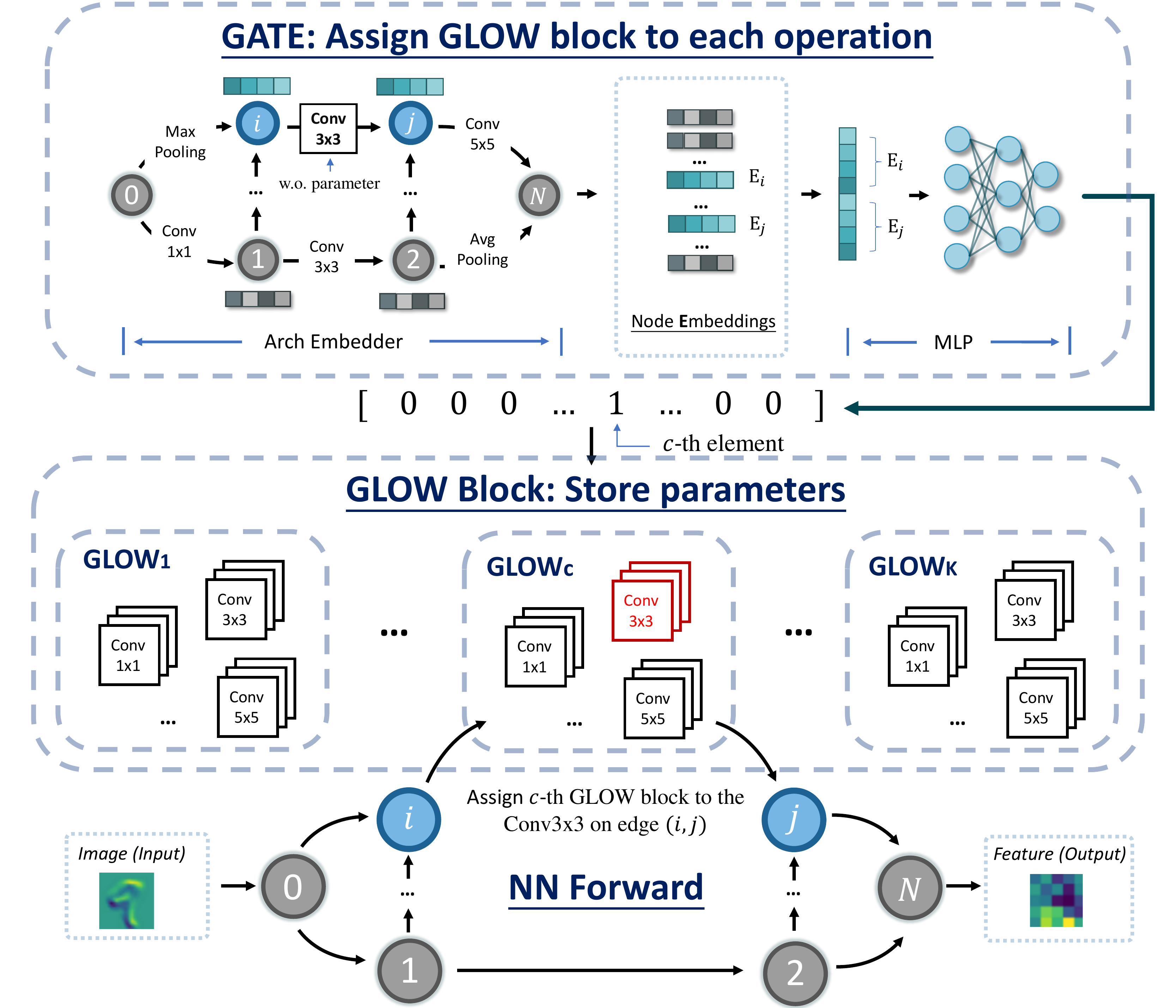}
    \caption{CLOSENet contains GLOW blocks to store the parameters of candidate operations, and the GATE module to assign GLOW blocks to operations.}
    \label{fig:closenet}
  \end{center}
\end{figure}

The design of CLOSENet is illustrated in Fig.~\ref{fig:closenet}.
The key idea behind CLOSENet is to decouple the parameters from operations to enable flexible sharing scheme and adjustable sharing extent. 
Specifically, we design the GLobal Operation Weight (GLOW) Block to store the parameters, and design a GATE module for assigning the proper GLOW block to each operation. 

For a better understanding, we take the generic cell-based topological search space 
as an example. We will show in the appendix that CLOSENet can also adapt to other types of search spaces, such as ResNet-like search spaces.

\subsubsection{Generic Cell-based Topological Search Space.}
In cell-based search space, a complete architecture is stacked by a cell-architecture for multiple times (e.g., 15 times on NAS-Bench-201). A cell-architecture can be represented as a \textit{directed acyclic graph (DAG)}. Each node $x^i$ represents a feature map, while each edge $o^{(i,j)}$ represents an operation that transforms $x^i$ to $x^j$ with the corresponding parameters. For each node $j$, the feature $x_j$ is defined as:
\begin{equation}
    x^j = \sum_{i<j} o^{(i,j)}(x^i, W^{(i,j)}),
    \label{eq:1}
\end{equation}
\noindent where $o(x, W)$ denotes that the operation $o$ transforms the feature $x$ with the parameters from $W$.

\subsubsection{GLobal Operation Weight (GLOW) Block.} 
The function of GLOW blocks is to store the parameters of candidate operations, as shown in Fig.~\ref{fig:closenet} (middle). 
In the forward pass (as shown in the bottom of Fig.~\ref{fig:closenet}), the GLOW blocks are assigned to each operation via the GATE module (will be introduced in the following), and each operation can use the parameters from its assigned block to process the input feature map.

Specifically, we denote $G_i$ as the $i$-$th$ GLOW block, and $c^{(i,j)}$ as the index of the assigned  block for the operation in edge $(i,j)$. Then, 
the computation of the feature $x^j$ in Eq.~\ref{eq:1} can be rewritten as:
\begin{equation}
    x^j = \sum_{i<j} o^{(i,j)}(x^i, G_{c^{(i,j)}})
    \label{eq:2}
\end{equation}


\subsubsection{GATE Module.} 
We design a GATE module for assigning GLOW blocks to operations. 
The GATE module consists of an architecture embedder and a MLP module, as shown in Fig.~\ref{fig:closenet} (top). 
We construct a GCN-based architecture embedder~\cite{ning2020generic}, and use it to compute the node embeddings in the architecture. 
Then, we concatenate the embeddings of the input and output nodes of each operation and feed it into the MLP to get the assignment of the GLOW block.

Specifically, we denote $E_i$ as the embedding of node $i$, and $K$ as the number of GLOW blocks. For a cell-architecture $a$ with $N$ nodes, we first obtain the node embeddings by the architecture embedder as Eq.~\ref{eq:3}, and then calculate the probability distribution in edge $(i,j)$ as Eq.~\ref{eq:4} and Eq.~\ref{eq:5}.

\begin{equation}
    [E_1, E_2, E_3, ..., E_N] = \text{ArchEmb}(a)
    \label{eq:3}
\end{equation}

\begin{equation}
    [\lambda_1^{(i,j)}, \lambda_2^{(i,j)}, \lambda_3^{(i,j)}, ..., \lambda_K^{(i,j)}] = \text{MLP}(concat(E_i, E_j))
    \label{eq:4}
\end{equation}

\begin{equation}
    \text{Pr}(c^{(i,j)}=k) = \frac{exp(\lambda_k^{(i,j)})}{\sum_{k'=1}^K exp(\lambda_{k'}^{(i,j)})}
    \label{eq:5}
\end{equation}

To allow the back-propagation of gradients, we apply the reparameterization trick on Eq.~\ref{eq:2} and Eq.~\ref{eq:5}, and rewrite the computation of $x^j$ as:

\begin{equation}
    x^j = \sum_{i<j} \sum_{k=1}^K h_{k}^{(i,j)} o^{(i,j)}(x^i, G_k),
    \label{eq:6}
\end{equation}

\begin{equation}
    h^{(i,j)} = \mathop{\arg\max}\limits_{k}(\lambda_k^{(i,j)}+g_k),
    \label{eq:7}
\end{equation}

\noindent where $h^{(i,j)}$ is a one-hot vector of dimension $K$, and $g_k$ are i.i.d samples from Gumbel(0, 1). To make Eq.~\ref{eq:7} differentiable, we relax the $\arg\max$ function to a softmax function as:

\begin{equation}
    \hat{h}_k^{(i,j)} = \frac{exp((\lambda_k^{(i,j)}+g_k)/\tau)}{\sum_{k'=1}^K exp((\lambda_{k'}^{(i,j)}+g_{k'})/\tau)},
    \label{eq:8}
\end{equation}

\noindent where $\tau$ is the Gumbel-Softmax temperature. We use Eq.~\ref{eq:7} in the forward pass, and use Eq.~\ref{eq:8} in the backward pass to allow gradient propagation.

\subsubsection{Adjustment of Sharing Extent} Denote $E^{(i,j)}$ as the set of cell-architectures that contain the edge from node $i$ to node $j$. For edge $(i,j)$, we define its sharing extent $s^{(i,j)}$ as the average number of architectures that share one GLOW block in CLOSENet. The sharing extent of the supernet $s$ equals the sum of sharing extent of all the edges:

\begin{equation}
    s = \sum_{i,j} s^{(i,j)} = \sum_{i,j} \frac{|E^{(i,j)}|}{K} = \frac{\sum_{i,j} |E^{(i,j)}|}{K}.
    \label{eq:9}
\end{equation}

\noindent Therefore, we can naturally adjust the sharing extent of CLOSENet by adding or reducing the GLOW blocks. CLOSENet with more GLOW blocks (larger $K$) has a smaller sharing extent and vice versa.

\begin{figure}[htb]
  \begin{center}
    \includegraphics[width=0.90\linewidth]{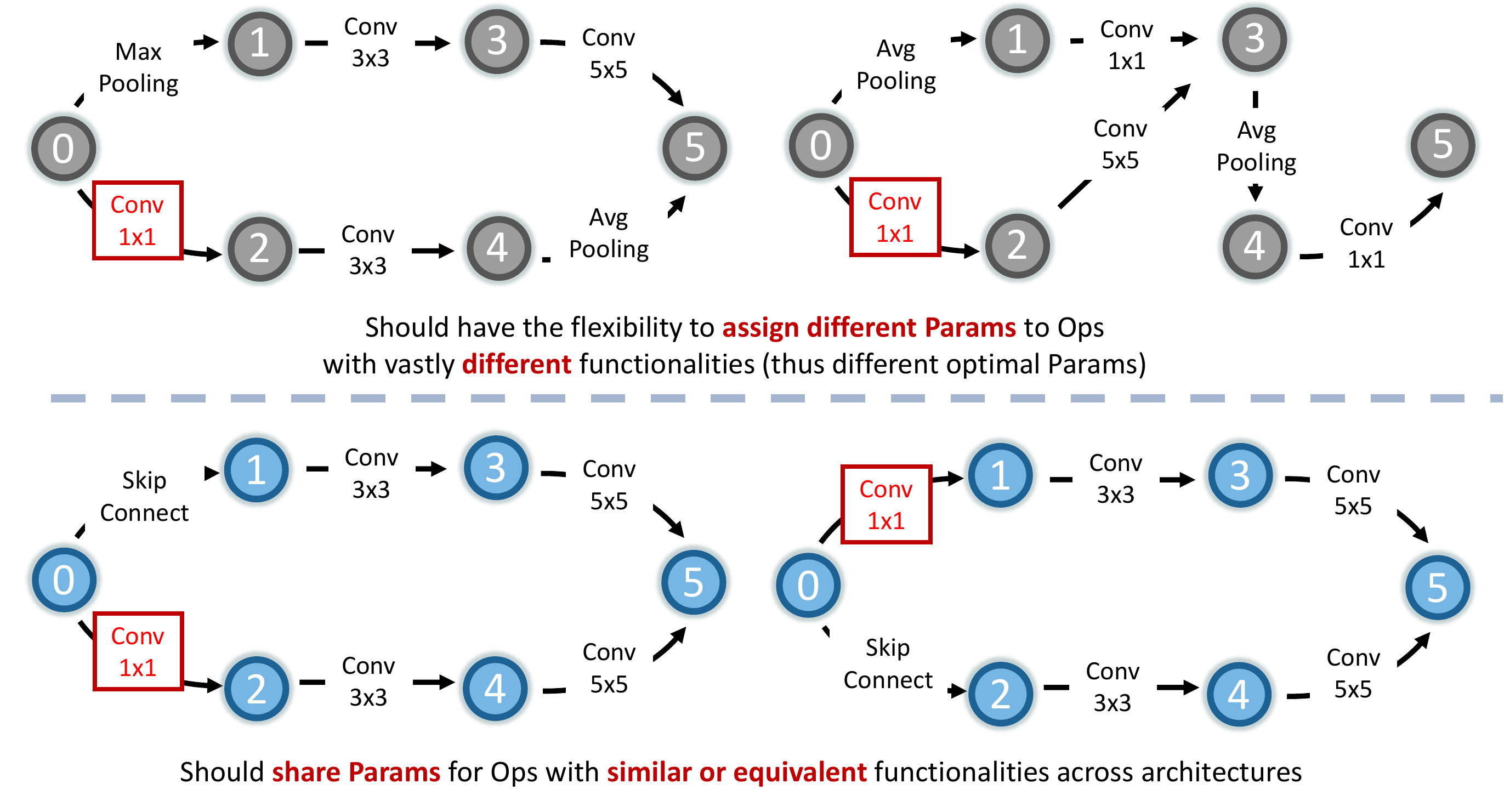}
    \caption{Two examples that show the strengths of CLOSENet. The sharing scheme of the vanilla supernet, i.e., sharing parameters between operations with the same position indexes, is improper in these two cases. In contrast, CLOSENet designates a more proper sharing pattern between operations according to their graph-based encoding given by GATE.}
    \label{fig:diff}
  \end{center}
\end{figure}

\subsubsection{Strengths Compared to Vanilla Supernets} 
The vanilla supernet and its variants (e.g., \textit{K-shot} and \textit{Few-shot} supernets~\cite{su2021k,zhao2021few}) preset the sharing scheme and extent by attaching a fixed set of parameters to each operation. On the contrary, CLOSENet decouples the parameters from the operations and enables \textbf{the dynamic decision of sharing scheme} based on a graph-based encoding of architecture operations. 
Specifically, the vanilla supernet shares parameters according to the position specified by the node indexes, i.e., the operations in the same ``position'' share the same parameters across different architectures. This sharing scheme is not flexible and can be suboptimal in some cases. For example, as shown in Fig.~\ref{fig:diff}(upper), the $1\times 1$ convolutions on the 0-2 edge share the same parameters between the two architectures, while they should have vastly different optimal parameters.
Intuitively, if two operations in two architectures have similar data processing functionality, it might be more reasonable to share their parameters. 
The design of CLOSENet matches this intuition:
The GATE module learns to pick the right GLOW block for each operation based on the graph-based encoding of all operations and topology in the cell architecture. Instead of presetting the sharing scheme according to the position information, CLOSENet takes a more flexible and reasonable way to dynamically determine which block each operation should use, and thereby designates which operations in different architectures should share their parameters.
For example, as shown in Fig.~\ref{fig:diff}(bottom), since the two $1\times 1$ convolutions are equivalent in two isomorphic architectures despite having different position indexes, it is reasonable for them to share parameters. The vanilla supernet uses different parameters for these two convolutions, while CLOSENet assigns the same GLOW block for them.





Moreover, this decoupling enables us to \textbf{flexibly adjust the sharing extent} by changing $K$ in Eq.~\ref{eq:9}. 
Thus, CLOSENet enables us to apply our curriculum learning-like training strategy. 
In summary, both the dynamic sharing scheme and the adjustable sharing extent make CLOSENet a more powerful supernet.

\subsection{CLOSE: Curriculum Learning On Sharing Extent}
\label{subsec:close}



We borrow the idea of curriculum learning to design a novel supernet training strategy CLOSE. 
Specifically, we initialize the CLOSENet with only one GLOW block at the beginning. This large sharing extent helps us to train the supernet much faster. 
Then, we gradually add GLOW blocks at preset epochs to reduce the sharing extent. 
In this way, CLOSE not only accelerates the supernet training, but also improves the saturating ranking quality of the supernet.

When switching the curriculum (i.e., increasing the sharing extent), we add a new GLOW block into CLOSENet and a corresponding MLP output unit to the GATE module. 
How to initialize the newly added parameters is critical to the performance of CLOSENet. 
Additionally, the regular schedule for the learning rate does not fit for CLOSE (see below). Correspondingly, we propose two techniques, the Weight Inherited Technique and the Schedule Restart Technique.

\begin{figure}[h]
  \begin{center}
    \includegraphics[width=0.85\linewidth]{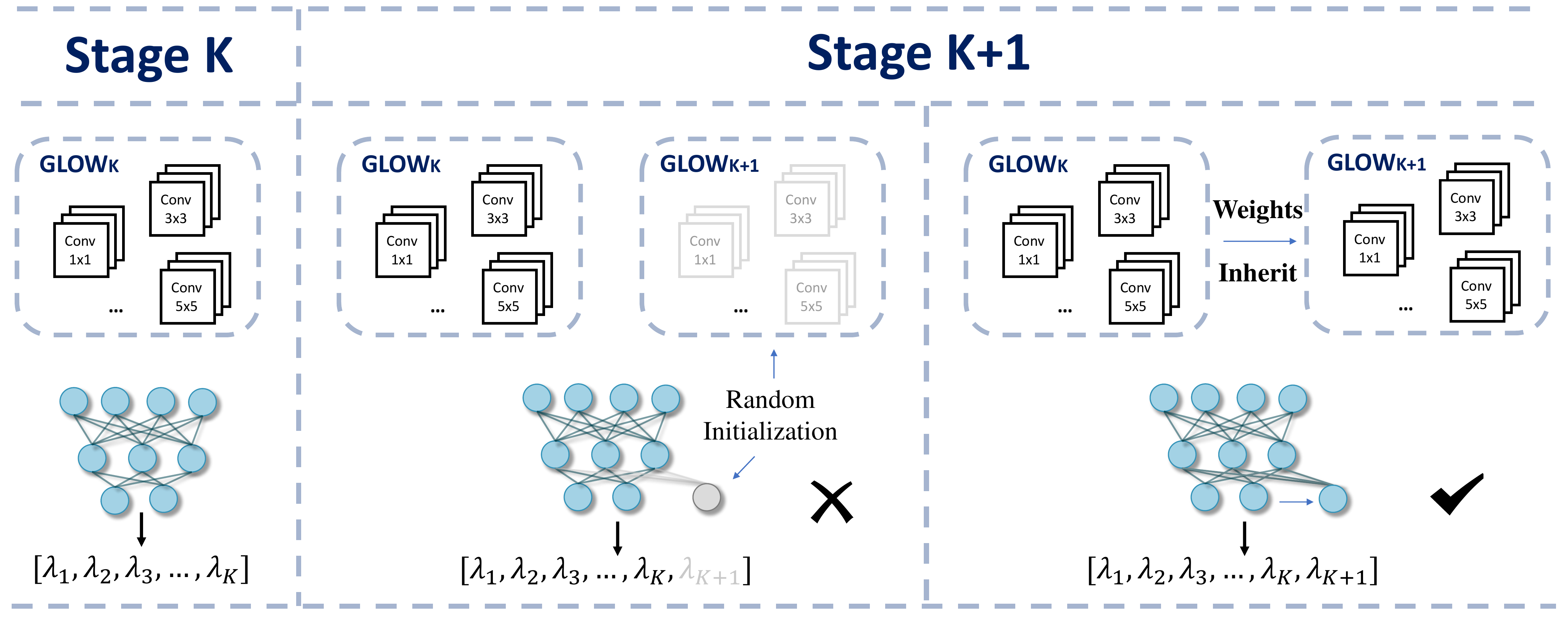}
    \caption{WIT for GLOW blocks (top) and the MLP output units (bottom).}
    \label{fig:wit}
  \end{center}
\end{figure}

\subsubsection{Weight Inherit Technique (WIT).} 
Instead of randomly initializing the new GLOW block and MLP output unit, we make their weights inherit from those of previous GLOW blocks and MLP output units,
as shown in Fig.~\ref{fig:wit}. 
This helps with the more efficient training of the new GLOW block and MLP unit.

\subsubsection{Schedule Restart Technique (SRT).}
In the training process, the learning rate is reduced gradually to approach the optimal solution. That is to say, it will become quite small after many epochs. However, following this schedule, CLOSE might fail to jump out of the local optimal solution of the preceding curriculum. 
To overcome this problem, we propose to restart the learning rate and schedule at preset epochs. With SRT, CLOSE can quickly reach the new optimal solution after switching to a new curriculum.



\section{Experiments}
\label{sec:exp}


\subsection{Evaluation of Ranking Quality}
\label{sec:evalranking}

We evaluate our method on four NAS search spaces, including NAS-Bench-201~\cite{dong2020bench}, NAS-Bench-301~\cite{siems2020bench}, NDS ResNet~\cite{radosavovic2020designing} and NDS ResNeXt-A~\cite{radosavovic2020designing}. The training configurations are shown in the appendix. 
Following previous studies~\cite{ning2020surgery,ning2020generic}, we use two evaluation criteria as follows: 

\begin{itemize}
\item Kendall's Tau (KD): The relative difference of the number of concordant pairs and discordant pairs, which reflects the overall ranking correlation.

\item Precision@topK (P@topK): The proportion of true top-K architectures in the top-K architectures according to the one-shot estimations, which reflects the ability of identifying the top-performing architectures.

\end{itemize}


\subsubsection{Comparison with Vanilla One-Shot Baselines} 
\begin{figure}[h]
    \centering
    \includegraphics[width=0.8\linewidth]{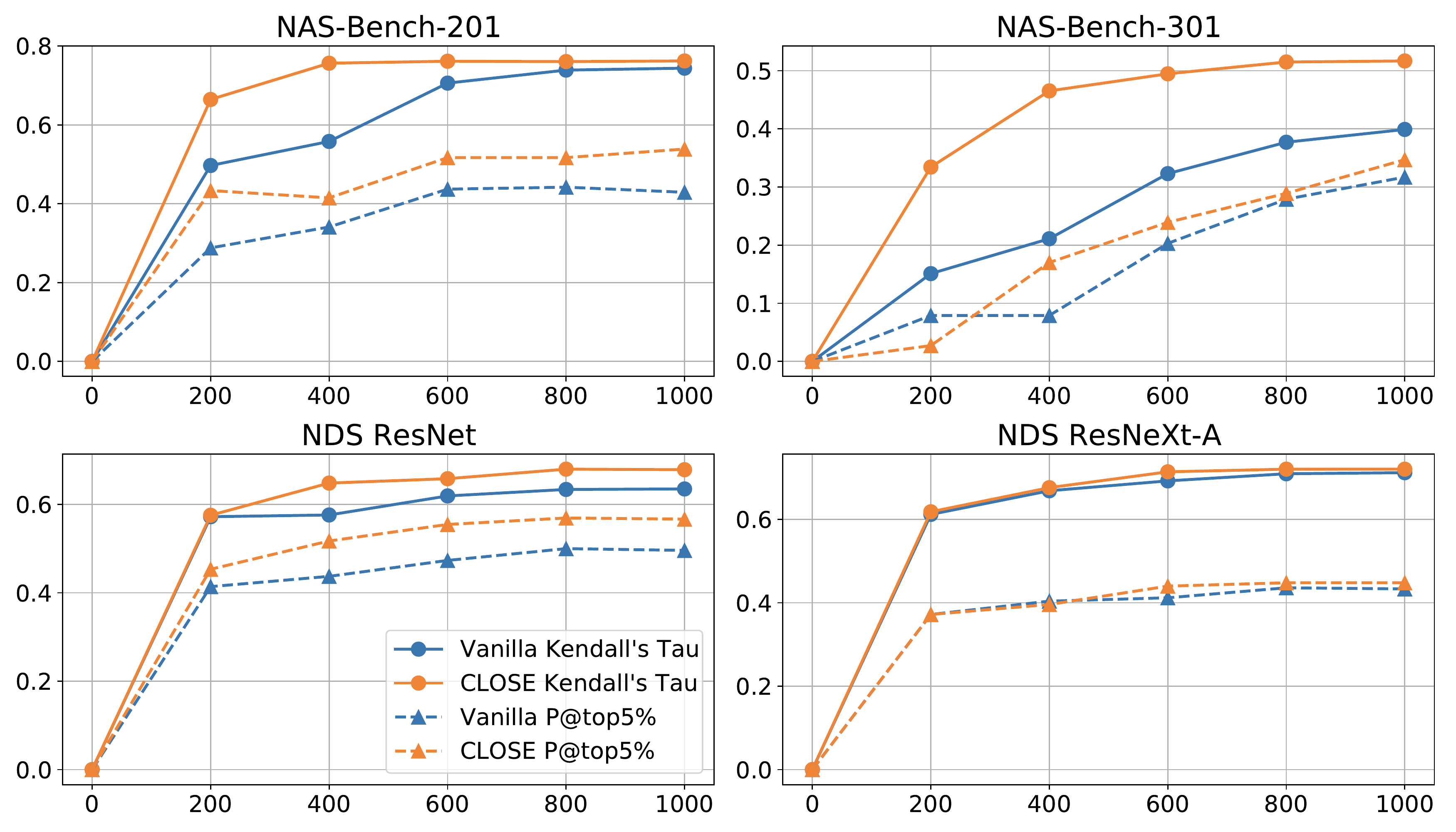}
    \caption{Comparison of different criteria with the vanilla one-shot supernet on four NAS benchmarks. X-axis: Training epochs. Y-axis: Evaluation criteria.}
    \label{fig:trend}
\end{figure}

We compare CLOSENet with vanilla supernets on four NAS benchmarks. As shown in Fig.~\ref{fig:trend}, CLOSENet achieves a higher KD and P@top5\% on all the NAS benchmarks. Moreover, we can see that throughout the training process, CLOSENet consistently achieves higher ranking quality, which implies CLOSENet's superiority to the vanilla supernet under any budget for supernet training.

\subsubsection{Comparison with Improved One-Shot Methods} 
\begin{figure}[h]
\begin{minipage}{0.47\linewidth}
\includegraphics[width=0.93\linewidth]{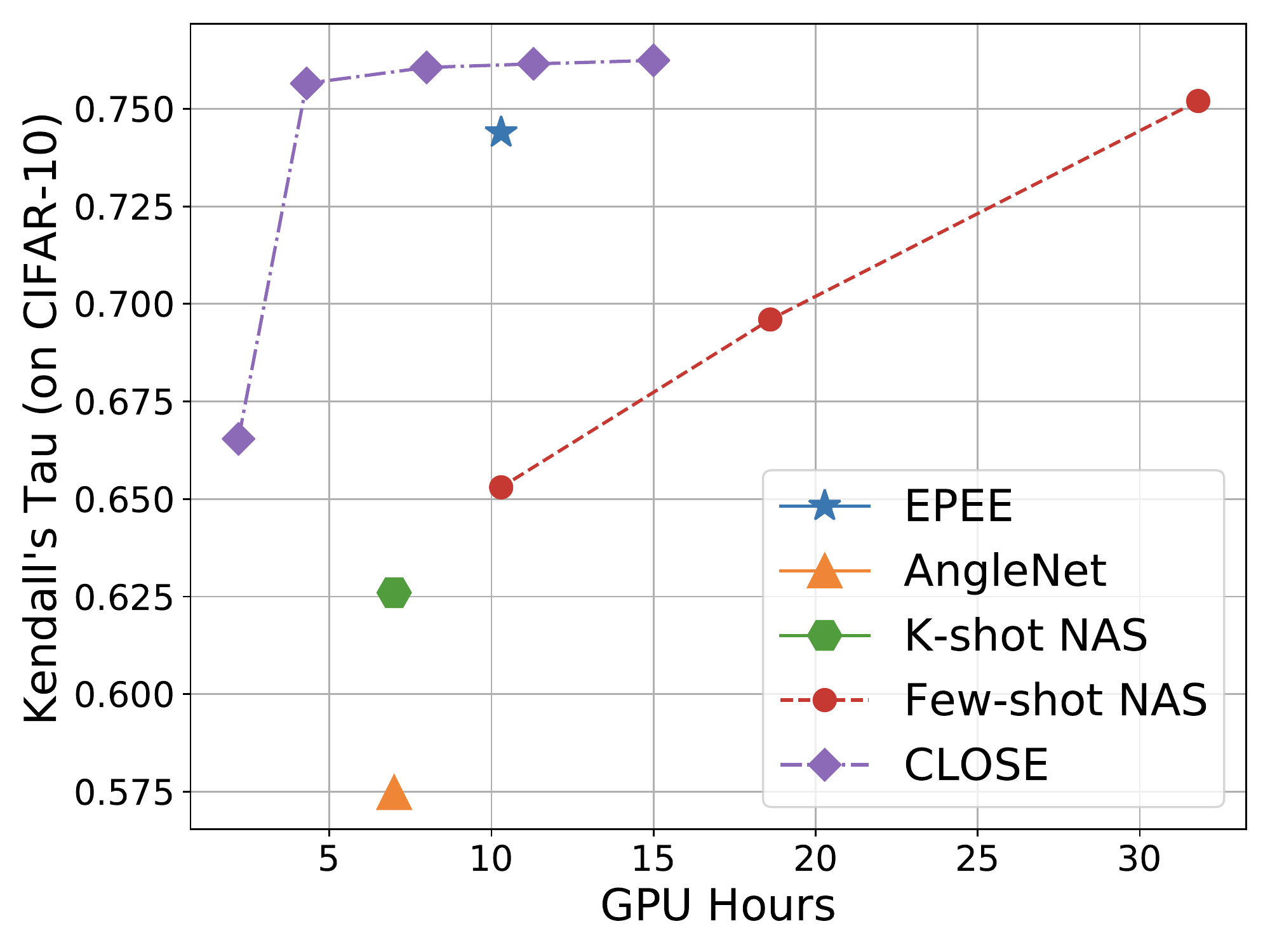}
\caption{Comparison with previous improved methods on NAS-Bench-201 for three datasets, i.e. CIFAR-10, CIFAR-100 (C-10) and ImageNet-16 (IN-16).}
\label{fig:nb201_kd}
\end{minipage}
\begin{minipage}[b]{0.45\linewidth}
\centering
\renewcommand\arraystretch{1.2}
\begin{tabular}{p{2.0cm}<{\centering} | p{1.5cm}<{\centering}  p{1.5cm}<{\centering}}
\toprule
\multirow{2}{*}{Method} & \multicolumn{2}{c}{Kendall's Tau} \\ \cmidrule(lr){2-3} & C-100 & IN-16 \\ \midrule
EPEE & 0.5600 & 0.5400  \\
AngleNet & 0.6040 & 0.5445   \\
K-shot NAS  & 0.6122 & 0.5633  \\
\cmidrule(lr){1-3}
\textbf{CLOSE} & \textbf{0.6693} & \textbf{0.6632} \\ \bottomrule
\end{tabular}
\end{minipage}
\end{figure}

Fig.~\ref{fig:nb201_kd} compares CLOSE with previous work on improving NAS evaluation strategy, including \textit{EPEE}~\cite{ning2020surgery}, \textit{AngleNet}~\cite{hu2020angle}, \textit{K-shot NAS}~\cite{su2021k} and \textit{Few-shot NAS}~\cite{zhao2021few}. Results show that CLOSE reaches SOTA KDs on all the three datasets of NAS-Bench-201.

\subsection{Evaluation of Search Performance}

We combine CLOSE with various search strategies, including 
DARTS~\cite{liu2018darts}, SNAS~\cite{xie2018snas} and CARS~\cite{yang2020cars}, to evaluate whether it improves the search performance. 

\subsubsection{Results}
\begin{figure}[h!]
    \centering
    \includegraphics[width=0.90\linewidth]{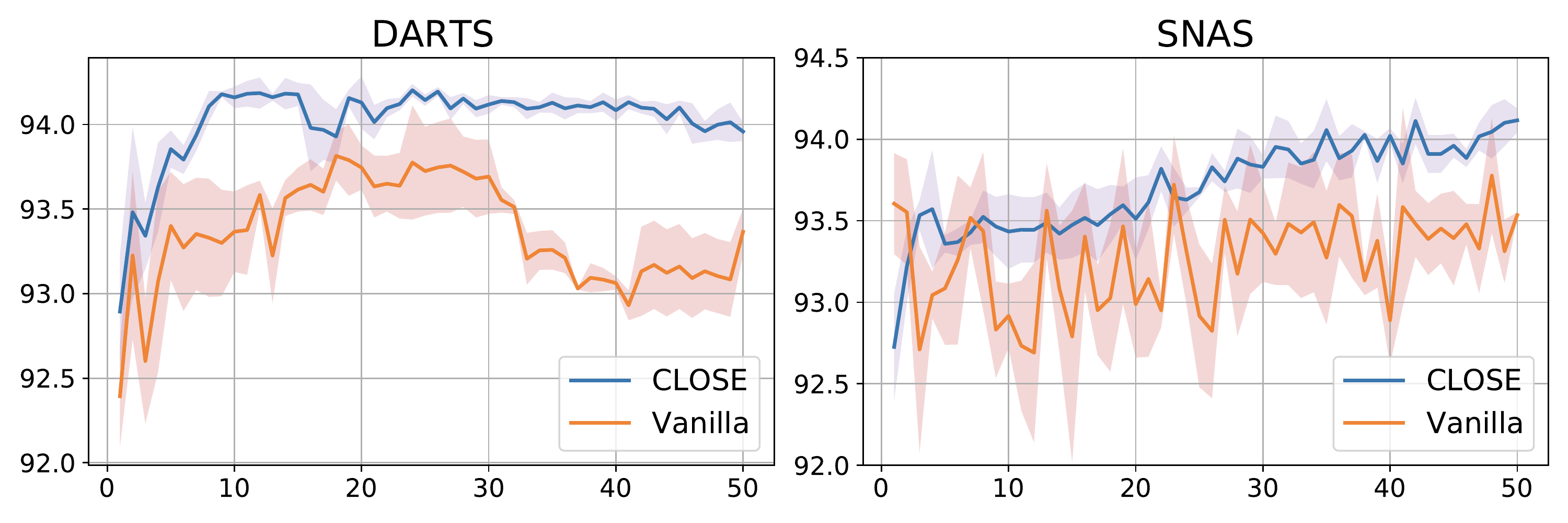}
    \caption{Evaluation of CLOSE with two search strategies in the NAS-Bench-301 search space. X-axis: Training epochs. Y-axis: Test accuracy.}
    \label{fig:nb301_search_performance}
\end{figure}

We run DARTS and SNAS search with CLOSE on NAS-Bench-301 and show the derived architecture accuracy in Fig.~\ref{fig:nb301_search_performance}. %
We can see that CLOSE benefits the search process significantly. In particular, it can alleviate the collapse issue of DARTS caused by the improper preference of parameter-free operations (i.e., \textit{skip\_connect}) in early training stages~\cite{liang2019darts+,hong2020dropnas}, as it provides a less biased estimation (see Appendix 1.1). 

We run CARS with CLOSE in the DARTS search space and 
Tab.~\ref{tab:cifar_imgnet} shows the performances of the discovered architecture. As can be seen, CLOSE achieves a competitive test error of 2.72\% in CIFAR-10. 
And when transferred to the ImageNet, the found architecture achieves a low test error of 24.7\%.
\begin{table*}[h]
\centering
\caption{Comparison of architecture performances on CIFAR-10 and ImageNet.}
\begin{tabular}{p{3.5cm}<{\centering}|p{2cm}<{\centering}cp{2cm}<{\centering}|p{2cm}<{\centering}c}
\toprule
\multirow{3}{*}{Method} & \multicolumn{3}{c}{CIFAR-10} & \multicolumn{2}{c}{ImageNet} \\ \cmidrule(lr){2-4} \cmidrule(lr){5-6}
& Top-1 Error & Param & Search Cost & Top-1 Error & Param \\
& (\%) & (M) & (GPU days) & (\%) & (M) \\
\midrule
NASNet-A~\cite{zoph2018learning}  & 2.65 & 3.3 & 2000 & 26.0 & 5.3 \\
AmoebaNet-B~\cite{real2019regularized} & 2.55 & 2.8&3150 & 26.0 & 5.3 \\
PNAS~\cite{liu2018progressive}  & 3.41 & 5.1& 225 & 25.8& 5.1 \\ \midrule
ENAS~\cite{pham2018efficient}  & 2.89  & 4.6  & 0.5 & - & - \\
DARTS~\cite{liu2018darts} & 2.76   & 3.3  & 1.5 & 26.9   & 4.9 \\
SNAS~\cite{xie2018snas} & 2.85   & 2.8    & 1.5 & 27.3   & 4.3 \\
BayesNAS~\cite{zhou2019bayesnas} & 2.81  & 3.4  &  0.2 & 26.5   & 3.9 \\
GDAS~\cite{dong2019searching} & 2.82 & 2.5 & 0.17 & 27.5 & 4.4 \\
\midrule
CLOSE (Ours)  & 2.72 $\pm$ 0.04  & 4.1  & 0.6 & 24.7  & 4.8 \\
\bottomrule
\end{tabular}
\label{tab:cifar_imgnet}
\end{table*}

\subsection{Ablation Studies}

\subsubsection{Effect of Number of Curriculums.} 
\begin{figure}[h]
    \centering
    \includegraphics [width=0.90\linewidth] {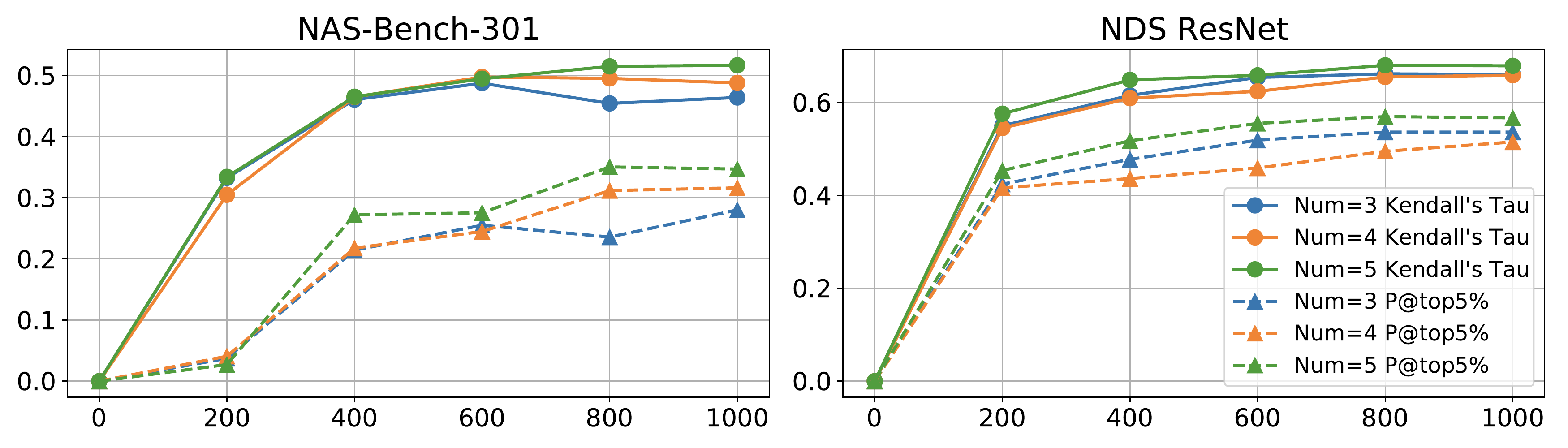}
    \caption{The ranking quality of CLOSE with different numbers of curriculums. X-axis: Training epochs. Y-axis: Values of criteria.}
    \label{fig:ablation_trend}
\end{figure}

We conduct an ablation study on the number of curriculums in two different types of search spaces, NAS-Bench-301 (a topological search space) and NDS ResNet (a non-topological search space). 
Results in Fig.~\ref{fig:ablation_trend} show that, in most cases, using more curriculums can improve the ranking quality of CLOSE. 
\subsubsection{Effect of WIT and SRT.} 
Tab.~\ref{tab:ablation} demonstrates the effect of WIT and SRT adopted by CLOSE on NAS-Bench-301 and NDS ResNet. Results show that these two techniques are both necessary for the ranking quality of CLOSE. 
\begin{table*}[h!]
\centering
\caption{The ranking quality of CLOSE w./w.o. the proposed techniques.}
\begin{tabular}{p{1.5cm}<{\centering}p{1.5cm}<{\centering}|p{2cm}<{\centering}p{2cm}<{\centering}|p{2cm}<{\centering}p{2cm}<{\centering}}
\toprule
\multirow{2}{*}{WIT} & \multirow{2}{*}{SRT} & \multicolumn{2}{c}{NAS-Bench-301} & \multicolumn{2}{c}{NDS ResNet} \\ \cmidrule(lr){3-4} \cmidrule(lr){5-6}
& & KD & P@top5\% & KD & P@top5\% \\
\midrule

 & & 0.1104 & 0.1145 & 0.6339 & 0.5387 \\
\checkmark & & 0.1047 & 0.1122 & 0.6550 & 0.5520 \\
 & \checkmark & 0.2004 & 0.1610 & 0.6448 & 0.5280 \\
 \checkmark & \checkmark & \textbf{0.5168} & \textbf{0.3470} & \textbf{0.6786} & \textbf{0.5667} \\

\bottomrule
\end{tabular}
\label{tab:ablation}
\end{table*}
\subsubsection{Effect of GATE Module.}
We compare the ranking quality of using the GATE module with randomly assigning GLOW blocks to operations. 
Results in Tab.~\ref{tab:ablation_gate} reveal that our learnable GATE module plays an essential role in CLOSENet.
\begin{table*}[h]
\centering
\caption{The ranking quality of CLOSE w./w.o. the GATE module}
\begin{tabular}{p{1.5cm}<{\centering}|p{2cm}<{\centering}p{2cm}<{\centering}|p{2cm}<{\centering}p{2cm}<{\centering}}
\toprule
\multirow{2}{*}{GATE} & \multicolumn{2}{c}{NAS-Bench-201} & \multicolumn{2}{c}{NAS-Bench-301} \\ \cmidrule(lr){2-3} \cmidrule(lr){4-5}
& KD & P@top5\% & KD & P@top5\% \\
\midrule
 w/o. & 0.3627 & 0.2014 & 0.2236 & 0.1924 \\
 w. & \textbf{0.7622} & \textbf{0.5387} & \textbf{0.5168} & \textbf{0.3470} \\
\bottomrule
\end{tabular}
\label{tab:ablation_gate}
\end{table*}
\subsubsection{Effect of Gradually Adding GLOW Blocks.} To show the benefit of gradually adding GLOW blocks, we conduct two contrast experiments. In the first experiment, we keep a fixed number of GLOW blocks in the training process. Results in Tab.~\ref{tab:fixed} demonstrate CLOSE performs better than fixing the sharing extent. In the second experiment, we gradually add blocks and stop after adding a certain number of blocks. Results in Tab.~\ref{tab:adding} show that the final ranking quality at 1000 epoch will degrade if GLOW blocks are not sufficiently added.

\begin{table}[h]
\caption{The ranking quality of supernets that use a \textit{fixed} number of blocks}
\centering
\begin{tabular}{p{2cm}<{\centering}|p{1.5cm}<{\centering}p{1.5cm}<{\centering}p{1.5cm}<{\centering}p{1.5cm}<{\centering}p{2cm}<{\centering}}
\toprule
\multirow{2}{*}{Benchmark} & \multicolumn{4}{c}{Fixed number of blocks} & \multirow{2}{*}{CLOSE} \\
\cmidrule(lr){2-5} & 2 & 3 & 4 & 5  \\ \midrule
NB201 & 0.7320 & 0.7247 & 0.7073 & - & \textbf{0.7622}   \\
NB301 & 0.4533 & 0.3427 & 0.3301 & 0.3106 & \textbf{0.5168} \\
\bottomrule
\end{tabular}
\label{tab:fixed}
\end{table}

\begin{table}[h!]
\caption{The ranking quality of supernets that add \textit{fewer} number of blocks.}
\centering
\begin{tabular}{p{2cm}<{\centering}|p{1.5cm}<{\centering}p{1.5cm}<{\centering}p{1.5cm}<{\centering}p{1.5cm}<{\centering}p{1.5cm}<{\centering}}
\toprule
\multirow{2}{*}{Benchmark} & \multicolumn{5}{c}{Number of added blocks in total}  \\
\cmidrule(lr){2-6} & 1 & 2 & 3 & 4 & 5  \\ \midrule
NB201 & 0.7050 & 0.7072 & 0.7502 & \textbf{0.7622} & -  \\
NB301 & 0.3990 & 0.4500 & 0.4641 & 0.4879 & \textbf{0.5168}   \\
\bottomrule
\end{tabular}
\label{tab:adding}
\end{table}

\section{Conclusions}
\label{sec:con}
This work borrows the idea of curriculum learning and proposes a novel training strategy CLOSE to train the NAS supernet both efficiently and effectively. Specifically, CLOSE adopts a curriculum learning-like schedule on the parameter sharing extent of supernets. 
To support this strategy, we design a novel one-shot supernet, namely CLOSENet, of which the sharing extent can be flexibly adjusted and the sharing scheme is decided based on a graph-based encoding. Extensive experiments demonstrate that equipped with CLOSENet, our proposed method CLOSE reaches a SOTA ranking quality on four NAS benchmarks. When searching in large search spaces, CLOSE can help to discover architectures with superior performances.


\section*{Acknowledgments}
This work was supported by National Natural Science Foundation of China (No. U19B2019, 61832007), National Key Research and Development Program of China (No. 2019YFF0301500),
Tsinghua EE Xilinx AI Research Fund, Beijing National Research Center for Information Science and Technology (BNRist), and Beijing Innovation Center for Future Chips.

\clearpage
%
%

\begin{thebibliography}{10}
\providecommand{\url}[1]{\texttt{#1}}
\providecommand{\urlprefix}{URL }
\providecommand{\doi}[1]{https://doi.org/#1}

\bibitem{bender2018understanding}
Bender, G., Kindermans, P.J., Zoph, B., Vasudevan, V., Le, Q.: Understanding
  and simplifying one-shot architecture search. In: International Conference on
  Machine Learning (ICML). pp. 550--559. PMLR (2018)

\bibitem{bengio2009curriculum}
Bengio, Y., Louradour, J., Collobert, R., Weston, J.: Curriculum learning. In:
  International Conference on Machine Learning (ICML). pp. 41--48 (2009)

\bibitem{benyahia2019overcoming}
Benyahia, Y., Yu, K., Smires, K.B., Jaggi, M., Davison, A.C., Salzmann, M.,
  Musat, C.: Overcoming multi-model forgetting. In: International Conference on
  Machine Learning (ICML). pp. 594--603. PMLR (2019)

\bibitem{brock2017smash}
Brock, A., Lim, T., Ritchie, J.M., Weston, N.: Smash: one-shot model
  architecture search through hypernetworks. In: International Conference on
  Learning Representations (ICLR) (2018)

\bibitem{dong2019searching}
Dong, X., Yang, Y.: Searching for a robust neural architecture in four gpu
  hours. In: IEEE Conference on Computer Vision and Pattern Recognition (CVPR).
  pp. 1761--1770 (2019)

\bibitem{dong2020bench}
Dong, X., Yang, Y.: Nas-bench-201: Extending the scope of reproducible neural
  architecture search. In: International Conference on Learning Representations
  (ICLR) (2020)

\bibitem{gong2019multi}
Gong, C., Yang, J., Tao, D.: Multi-modal curriculum learning over graphs. ACM
  Transactions on Intelligent Systems and Technology (TIST)  \textbf{10}(4),
  1--25 (2019)

\bibitem{guo2018curriculumnet}
Guo, S., Huang, W., Zhang, H., Zhuang, C., Dong, D., Scott, M.R., Huang, D.:
  Curriculumnet: Weakly supervised learning from large-scale web images. In:
  European Conference on Computer Vision (ECCV). pp. 135--150 (2018)

\bibitem{guo2020breaking}
Guo, Y., Chen, Y., Zheng, Y., Zhao, P., Chen, J., Huang, J., Tan, M.: Breaking
  the curse of space explosion: Towards efficient nas with curriculum search.
  In: International Conference on Machine Learning (ICML). pp. 3822--3831. PMLR
  (2020)

\bibitem{guo2020single}
Guo, Z., Zhang, X., Mu, H., Heng, W., Liu, Z., Wei, Y., Sun, J.: Single path
  one-shot neural architecture search with uniform sampling. In: European
  Conference on Computer Vision (ECCV). pp. 544--560. Springer (2020)

\bibitem{he2016deep}
He, K., Zhang, X., Ren, S., Sun, J.: Deep residual learning for image
  recognition. In: IEEE Conference on Computer Vision and Pattern Recognition
  (CVPR). pp. 770--778 (2016)

\bibitem{hong2020dropnas}
Hong, W., Li, G., Zhang, W., Tang, R., Wang, Y., Li, Z., Yu, Y.: Dropnas:
  Grouped operation dropout for differentiable architecture search. In:
  International Joint Conference on Artificial Intelligence (IJCAI). pp.
  2326--2332 (2020)

\bibitem{hu2020angle}
Hu, Y., Liang, Y., Guo, Z., Wan, R., Zhang, X., Wei, Y., Gu, Q., Sun, J.:
  Angle-based search space shrinking for neural architecture search. In:
  European Conference on Computer Vision (ECCV). pp. 119--134. Springer (2020)

\bibitem{jiang2014easy}
Jiang, L., Meng, D., Mitamura, T., Hauptmann, A.G.: Easy samples first:
  Self-paced reranking for zero-example multimedia search. In: ACM
  International Multimedia Conference (MM). pp. 547--556 (2014)

\bibitem{karras2017progressive}
Karras, T., Aila, T., Laine, S., Lehtinen, J.: Progressive growing of gans for
  improved quality, stability, and variation. In: International Conference on
  Learning Representations (ICLR). OpenReview.net (2018)

\bibitem{liang2019darts+}
Liang, H., Zhang, S., Sun, J., He, X., Huang, W., Zhuang, K., Li, Z.: Darts+:
  Improved differentiable architecture search with early stopping. arXiv
  preprint arXiv:1909.06035  (2019)

\bibitem{liu2018progressive}
Liu, C., Zoph, B., Neumann, M., Shlens, J., Hua, W., Li, L.J., Fei-Fei, L.,
  Yuille, A., Huang, J., Murphy, K.: Progressive neural architecture search.
  In: European Conference on Computer Vision (ECCV). pp. 19--34 (2018)

\bibitem{liu2018darts}
Liu, H., Simonyan, K., Yang, Y.: Darts: Differentiable architecture search. In:
  International Conference on Learning Representations (ICLR) (2019)

\bibitem{luo2019understanding}
Luo, R., Qin, T., Chen, E.: Understanding and improving one-shot neural
  architecture optimization. CoRR  \textbf{abs/1909.10815} (2019)

\bibitem{ning2020surgery}
Ning, X., Tang, C., Li, W., Zhou, Z., Liang, S., Yang, H., Wang, Y.: Evaluating
  efficient performance estimators of neural architectures. In: Annual
  Conference on Neural Information Processing Systems (NIPS) (2021)

\bibitem{ning2020generic}
Ning, X., Zheng, Y., Zhao, T., Wang, Y., Yang, H.: A generic graph-based neural
  architecture encoding scheme for predictor-based nas. In: European Conference
  on Computer Vision (ECCV). pp. 189--204. Springer (2020)

\bibitem{niu2020disturbance}
Niu, S., Wu, J., Zhang, Y., Guo, Y., Zhao, P., Huang, J., Tan, M.:
  Disturbance-immune weight sharing for neural architecture search. Neural
  Networks  \textbf{144},  553--564 (2021)

\bibitem{pham2018efficient}
Pham, H., Guan, M., Zoph, B., Le, Q., Dean, J.: Efficient neural architecture
  search via parameters sharing. In: International Conference on Machine
  Learning (ICML). pp. 4095--4104. PMLR (2018)

\bibitem{platanios2019competence}
Platanios, E.A., Stretcu, O., Neubig, G., P{\'o}czos, B., Mitchell, T.:
  Competence-based curriculum learning for neural machine translation. In:
  Proceedings of the 2019 Conference of the North American Chapter of the
  Association for Computational Linguistics: Human Language Technologies,
  Volume 1 (Long and Short Papers). pp. 1162--1172 (2019)

\bibitem{radosavovic2020designing}
Radosavovic, I., Kosaraju, R.P., Girshick, R., He, K., Doll{\'a}r, P.:
  Designing network design spaces. In: IEEE Conference on Computer Vision and
  Pattern Recognition (CVPR). pp. 10428--10436 (2020)

\bibitem{real2019regularized}
Real, E., Aggarwal, A., Huang, Y., Le, Q.V.: Regularized evolution for image
  classifier architecture search. In: AAAI Conference on Artificial
  Intelligence. vol.~33, pp. 4780--4789 (2019)

\bibitem{ren2018self}
Ren, Z., Dong, D., Li, H., Chen, C.: Self-paced prioritized curriculum learning
  with coverage penalty in deep reinforcement learning. IEEE transactions on
  neural networks and learning systems  \textbf{29}(6),  2216--2226 (2018)

\bibitem{siems2020bench}
Siems, J., Zimmer, L., Zela, A., Lukasik, J., Keuper, M., Hutter, F.:
  Nas-bench-301 and the case for surrogate benchmarks for neural architecture
  search. arXiv preprint arXiv:2008.09777  (2020)

\bibitem{soviany2022curriculum}
Soviany, P., Ionescu, R.T., Rota, P., Sebe, N.: Curriculum learning: A survey.
  International Journal of Computer Vision (IJCV) pp. 1--40 (2022)

\bibitem{su2021k}
Su, X., You, S., Zheng, M., Wang, F., Qian, C., Zhang, C., Xu, C.: K-shot nas:
  Learnable weight-sharing for nas with k-shot supernets. In: International
  Conference on Machine Learning (ICML). pp. 9880--9890. PMLR (2021)

\bibitem{tay2019simple}
Tay, Y., Wang, S., Luu, A.T., Fu, J., Phan, M.C., Yuan, X., Rao, J., Hui, S.C.,
  Zhang, A.: Simple and effective curriculum pointer-generator networks for
  reading comprehension over long narratives. In: Proceedings of the 57th
  Annual Meeting of the Association for Computational Linguistics. pp.
  4922--4931 (2019)

\bibitem{xie2017aggregated}
Xie, S., Girshick, R., Doll{\'a}r, P., Tu, Z., He, K.: Aggregated residual
  transformations for deep neural networks. In: IEEE Conference on Computer
  Vision and Pattern Recognition (CVPR). pp. 1492--1500 (2017)

\bibitem{xie2018snas}
Xie, S., Zheng, H., Liu, C., Lin, L.: Snas: stochastic neural architecture
  search. In: International Conference on Learning Representations (ICLR)
  (2019)

\bibitem{yang2020cars}
Yang, Z., Wang, Y., Chen, X., Shi, B., Xu, C., Xu, C., Tian, Q., Xu, C.: Cars:
  Continuous evolution for efficient neural architecture search. In: IEEE
  Conference on Computer Vision and Pattern Recognition (CVPR). pp. 1829--1838
  (2020)

\bibitem{sciuto2019evaluating}
Yu, K., Sciuto, C., Jaggi, M., Musat, C., Salzmann, M.: Evaluating the search
  phase of neural architecture search. In: International Conference on Learning
  Representations (ICLR) (2020)

\bibitem{zela2020bench}
Zela, A., Siems, J., Hutter, F.: Nas-bench-1shot1: Benchmarking and dissecting
  one-shot neural architecture search. In: International Conference on Learning
  Representations (ICLR) (2019)

\bibitem{zhang2020overcoming}
Zhang, M., Li, H., Pan, S., Chang, X., Su, S.: Overcoming multi-model
  forgetting in one-shot nas with diversity maximization. In: IEEE Conference
  on Computer Vision and Pattern Recognition (CVPR). pp. 7809--7818 (2020)

\bibitem{zhao2021few}
Zhao, Y., Wang, L., Tian, Y., Fonseca, R., Guo, T.: Few-shot neural
  architecture search. In: International Conference on Machine Learning (ICML).
  pp. 12707--12718. PMLR (2021)

\bibitem{zhou2019bayesnas}
Zhou, H., Yang, M., Wang, J., Pan, W.: Bayesnas: A bayesian approach for neural
  architecture search. In: International Conference on Machine Learning (ICML).
  pp. 7603--7613. PMLR (2019)

\bibitem{zoph2016neural}
Zoph, B., Le, Q.V.: Neural architecture search with reinforcement learning. In:
  International Conference on Learning Representations (ICLR) (2017)

\bibitem{zoph2018learning}
Zoph, B., Vasudevan, V., Shlens, J., Le, Q.V.: Learning transferable
  architectures for scalable image recognition. In: IEEE Conference on Computer
  Vision and Pattern Recognition (CVPR). pp. 8697--8710 (2018)

\end{thebibliography}

\clearpage
\appendix
\renewcommand{\thefigure}{A\arabic{figure}}
\renewcommand{\thetable}{A\arabic{table}}
\renewcommand{\theequation}{A\arabic{equation}}
\setcounter{figure}{0}
\setcounter{table}{0}

\section{Additional Discussions on CLOSE and CLOSENet}

\subsection{Insights into the Improvements by Increasing Sharing Extent}

\begin{figure}[htb]
  \begin{center}
    \includegraphics[width=0.8\linewidth]{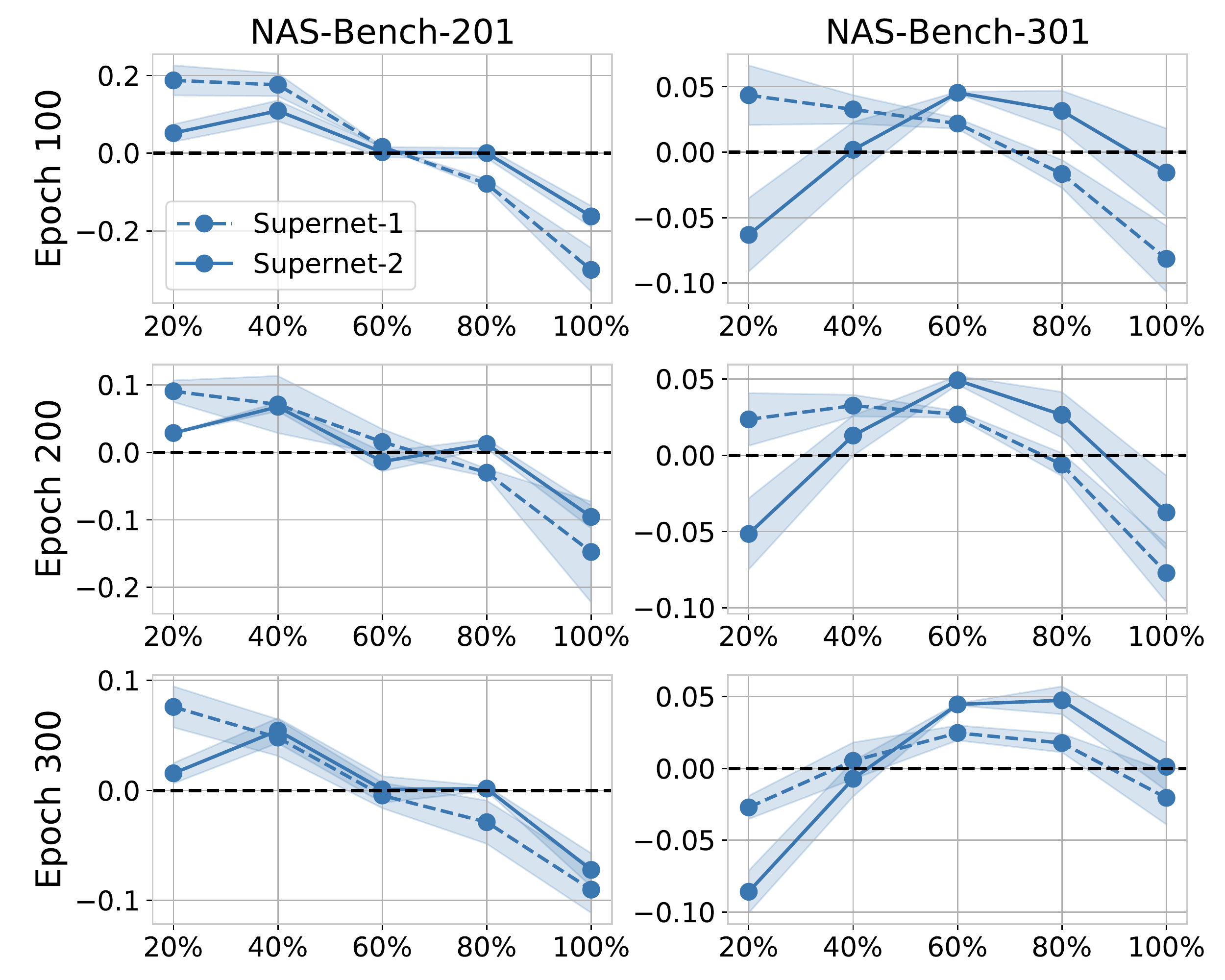}
    \caption{Evaluation of the under-estimation phenomenon on NAS-Bench-201 and NAS-Bench-301. X-axis: Complexity groups. Y-axis: Average RD.}
    \label{fig:overestimation}
  \end{center}
\end{figure}

Sec.~3.1 of the main paper shows that Supernet-2 (with larger sharing extent) significantly outperforms Supernet-1 (with vanilla sharing extent) in the early training stages. This observation is a bit counter-intuitive, since the previous studies have shown that a large sharing extent would aggravate the parameter coupling and multi-model forgetting phenomenon~\cite{benyahia2019overcoming,zhang2020overcoming,niu2020disturbance,ning2020surgery}. Therefore, it might be confusing where these improvements come from. To further understand our observation, we conduct a deeper inspection into the estimation results of the Supernet-1 and Supernet-2, and find out that there are two reasons underlying the observation.
\\
\\
\textbf{First, a larger parameter sharing extent accelerates the training process of the supernet}. 
This is due to the reduced number of the supernet's parameters. 
Previous studies reveal that a longer training can improve the ranking quality, yet brings extra computational costs~\cite{luo2019understanding,ning2020surgery}. In our case, the highly-shared supernet (Supernet-2) can achieve the same training performance under a smaller computational budget, thus improve the ranking quality.
\\
\\
\textbf{Second, a large parameter sharing extent alleviates the well-known under-estimation phenomenon of larger architectures in vanilla one-shot estimations~\cite{luo2019understanding,ning2020surgery}}. Following the visualization technique of a recent study~\cite{ning2020surgery}, we divide the candidate architectures into five groups based on their complexities, and obtain the average ranking difference of architectures in each group. The ranking difference (RD) of an architecture is defined as the difference of its true ranking and its estimated ranking by supernets. A negative RD indicates the under-estimation of an architecture. As shown in Fig.~\ref{fig:overestimation}, the average RDs of the largest architectures in the top 20\% (80\% and 100\% on the X-axis) 
in Supernet-2 are closer to zero than that in Supernet-1 on both NAS-Bench-201 and NAS-Bench-301. This indicates the alleviation of the under-estimation phenomenon of larger architectures. 

\subsection{More Discussion on Using The GATE Module for The Dynamic Decision of Sharing Scheme}

The ``Strengths compared to Vanilla Supernets'' section and Fig.~3 of the main paper have discussed how the dynamic decision of sharing scheme between architectures in CLOSENet can facilitate a more proper parameter sharing scheme between architectures. For example, recall that the two
1$\times$1 convolutions in Fig. 3 (bottom) are equivalent in two isomorphic architectures despite having
different positions. And the vanilla
supernet uses different parameters for these two convolutions, while CLOSENet gives out a more proper sharing scheme to share the same GLOW block for them. 

This strength comes from the dynamic decision design of the sharing scheme, and also partly from the GCN-based architecture embedder~\cite{ning2020generic} adopted in our GATE module.
As this architecture embedder conducts permutation-invariant aggregations on the graph,
it can naturally map the counterpart nodes in isomorphic architectures to the same node embeddings. Therefore, for the equivalent operations (i.e., the operations between the counterpart nodes) in two architectures, the MLP in the GATE module takes the same embeddings as input, and thus give out the same assignment.

The GCN-based architecture embedder adopted by CLOSENet, GATES~\cite{ning2020generic}, mimics the actual data processing to model the NN architecture. Corresponding to the computation flow of the cell-architecture in Sec.~3.2 of the main paper, the embedding of node $j$ (denoted as $E_j$ in Eq.~3 of the main paper) is defined as:
\begin{equation}
    E_j = \sum_{i<j} \sigma(\mbox{OpEmb}(o^{(i,j)})W_o) \odot E_iW_x, 
\end{equation}
where $\sigma$ is a sigmoid function, $\mbox{OpEmb}(o)$ gives out the embedding of this type of operation, and $W_o$ and $W_x$ denote two different linear transformation matrices. The embedding of the input node is randomly initialized.

\subsection{CLOSENet in Non-topological Search Spaces}
In non-topological search spaces (e.g., ResNet-like search space~\cite{radosavovic2020designing}), the computation blocks (i.e., operations) are put in sequential order, which is different from the topological structure in the generic search spaces. Therefore, we replace the normal GATE module with a simple but effective strategy that \textbf{assigns each GLOW block to some consecutive operations in an interval}. This assignment strategy comes from the intuition that the consecutive operations in the architectures have similar data processing functionality. Based on our analysis in Sec.~3.2 of the main paper, it is more reasonable to share the parameters of these operations. Since each operation has only one assigned block at the same time, the assignment intervals of the GLOW blocks are nonoverlapping.

When adding a new GLOW block, we propose to choose an existing block and divide its assignment interval down the middle. Then we assign the operations in one of the divided intervals to the new block. In this way, we can naturally apply the WIT to the new block to inherit the weights from the chosen one.

Fig.~5 in Sec.~4.1 of the main paper demonstrates that CLOSE consistently achieves higher ranking quality across the training process on the two non-topological search spaces (i.e., NDS ResNet and NDS ResNeXt-A).

\subsection{Implementation of CLOSE}


Alg.~\ref{alg:close} shows the pipeline of using CLOSE to train CLOSENet.  
Specifically, we construct CLOSENet with one GLOW block at the beginning. Then, we gradually add blocks to reduce the sharing extent at the preset epochs. WIT is applied to initialize the parameters of the new block and MLP unit. In each training iteration, we randomly sample an architecture to update the parameters of CLOSENet, including GLOW blocks and the GATE module, through the Eq.~3 to Eq.~8 introduced in Sec.~3.2 of the main paper. SRT is applied to restart the learning rate when it becomes too small.

\begin{algorithm*}[htb]
 \renewcommand{\algorithmicensure}{\textbf{Output:}} 
 \caption{The Training Process of CLOSE on CLOSENet}
 \label{alg:close}
 \begin{algorithmic}[1]
 
  \renewcommand{\algorithmicrequire}{\textbf{Input:}}
  \REQUIRE
  ~\\$D$: Training data; $T$: Training epochs; $A$: Architecture search space;
  ~\\$S_{CL}$: The set of switch points (epochs) of sharing extent
  ~\\$S_{LR}$: The set of restart points (epochs) of learning rate
   
   
  \renewcommand{\algorithmicrequire}{\textbf{Training Process:}}
  \REQUIRE
  \STATE Construct a randomly initialized CLOSENet $N_A$ with one GLOW block
  \FOR{$t = 1, \cdots, T$ }
  \IF {$t \in S_{CL}$}
  {
        \STATE Adding a new GLOW block in $N_A$
        \STATE Using WIT to initialize the new block and GATE module
  }
  \ENDIF
  \IF {$t \in S_{LR}$}
  {
        \STATE Using SRT to restart the learning rate and schedule
  }
  \ENDIF
  \FOR{$i = 1, \cdots, I$ }
  \STATE Randomly sample an architecture $a \in A$
  \STATE Sample a batch of training data from $D$
  \STATE Update parameters in $N_A$ 
  with Eq.~3 $\sim$ Eq.~8
  \ENDFOR
  \ENDFOR
  \ENSURE The well-trained CLOSENet $N_A$
 \end{algorithmic}
\end{algorithm*}

\section{Detailed Configurations}

\subsection{Supernet Training}

In our experiments, we use the same training configurations for vanilla one-shot supernets and CLOSENet. In detail, we train supernets via a SGD optimizer with momentum 0.9 and weight decay 5e-4 . The learning rate is set to 0.05 initially and decayed by 0.5 each time the supernet accuracy stops to increase for 30 epochs. In the training process, the dropout rate is set to 0.1, and the gradient norm is clipped to be less than 5. The batch size is set to 512. For each batch of examples, we randomly sample one architecture to update supernets' parameters. Besides, we set the training epochs of all supernets to 1000 epochs.

\subsection{Architecture Search}

\begin{figure*}[htb]
    \centering
    \subfigure[Normal Cell]{
    \includegraphics[width=0.65\linewidth]{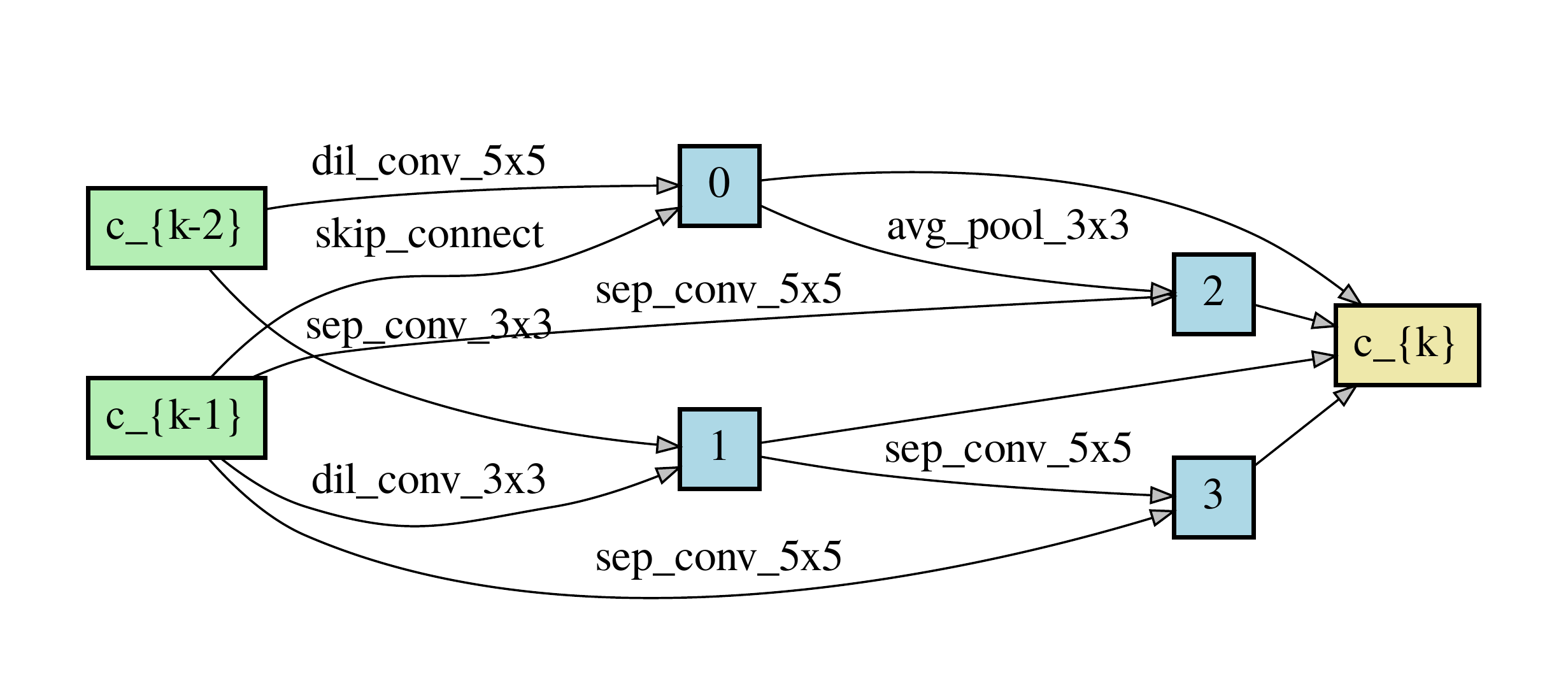}
    }
    \subfigure[Reduction Cell]{
    \includegraphics[width=0.8\linewidth]{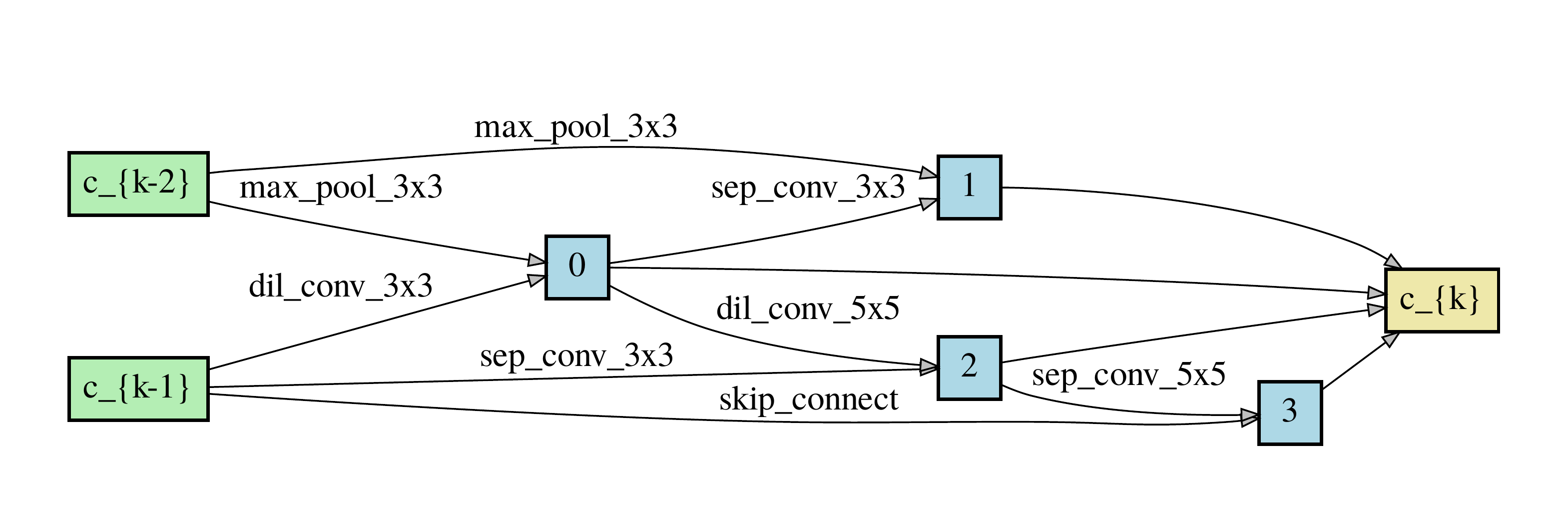}
    }
    \caption{The discovered cell architectures by CLOSE.} 
    \label{fig:best}
\end{figure*}

We adopt CARS~\cite{yang2020cars}, an improved evolutionary approach, to search the optimal architectures in DARTS search space. The search process contains the supernet training stage and the architecture search stage. We set the total epochs to 400, and set the population size to 100. We first train the supernet for 100 epochs to warm up the parameters.
Then, in the supernet training stage, we train the supernet for 5 epochs during one evolution iteration. 
In each mutation step during the architecture search stage,
random mutation, random crossover and random sampling are conducted with a probability of 0.25, 0.25, and 0.5, respectively, following CARS~\cite{yang2020cars}. Fig.~\ref{fig:best} shows the discovered architectures.

\subsection{Training of the Discovered Architectures}

On CIFAR-10, we stack the discovered architectures 20 times to construct the network, and set its initial channel number to 36. The network is trained for 600 epochs with batch size 128. We use a SGD optimizer with momentum 0.9 and weight decay 3e-4. The learning rate is decayed from 0.05 to 0.001 following a cosine schedule. The dropout rate is set to 0.1, and the gradient norm is clipped to be less than 5. Besides, the cutout augmentation with length 16, the path dropout of probability 0.2 and the auxiliary towers with weight 0.4 are used.

When transferring the discovered architectures to ImageNet, we stack 14 cells to construct the network, and set its initial channel number to 48. The network is trained for 300 epochs with batch size 256. The weight decay is set to 3e-5, and the learning rate is decayed from 0.1 to 0 following a cosine schedule. The path dropout technique is not used.

\section{Additional Experiments}










\subsection{Investigation of The GATE Module}

\begin{figure}[htb]
  \begin{center}
    \includegraphics[width=0.98\linewidth]{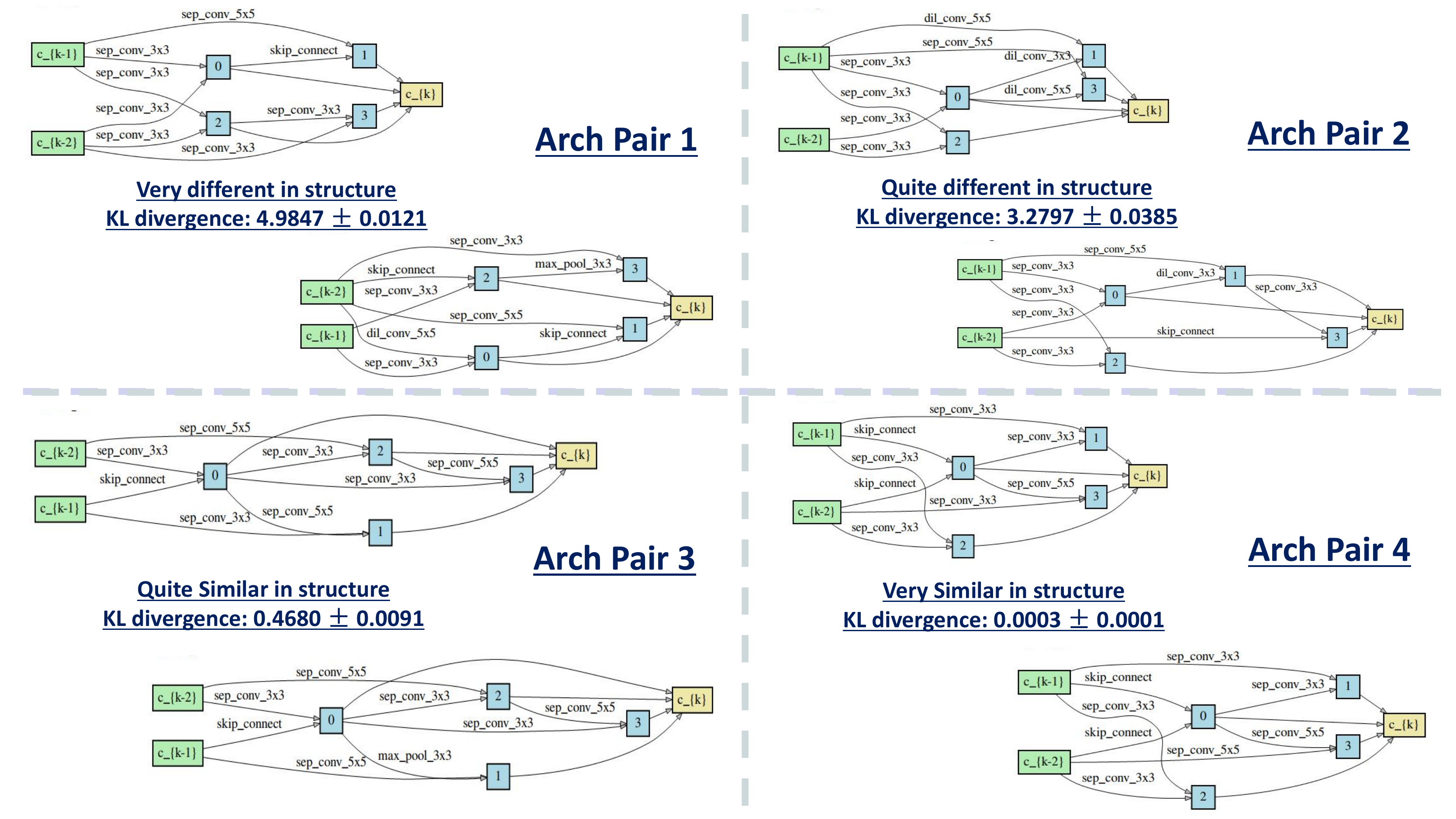}
    \caption{The four pairs of the architectures we sample to investigate the GATE module on NAS-Bench-301. We demonstrate the normal cells of the architectures here. Architectures in Arch Pair 1 and Arch Pair 2 are different in structure, while architectures in Arch Pair 3 and Arch Pair 4 are similar.}
    \label{fig:explain}
  \end{center}
\end{figure}

In this section, we conduct an experiment to further investigate the effectiveness and robustness of the GATE module in CLOSENet. Specifically, we randomly sample four pairs of architectures on NAS-Bench-301, where two of them have a big structure difference (labeled as 1 and 2), and the other two pairs (labeled as 3 and 4) have a similar structure. Fig.~\ref{fig:explain} shows the four pairs of architectures. 
We use a well-trained GATE module to obtain 
their assignment distribution following the Eq.~4 of the main paper. Then we calculate the KL divergence of the distribution between all the $3\times 3$ convolutions in each pair of architectures and make a summation, which can reflect their assignment similarity. 
The results of these four pairs are $4.9847\pm0.0121$, $3.2797\pm0.0385$, $0.4680\pm0.0091$ and $0.0003\pm0.0001$. The stability of the KL divergence across different random seeds shows the robustness of GATE. Meanwhile, the larger KL divergence of pair 1 and 2 also demonstrates the GATE module can give out a proper assignment of blocks.

\subsection{Effectiveness of The WIT Technique}

\begin{figure}[htb]
  \begin{center}
    \includegraphics[width=0.9\linewidth]{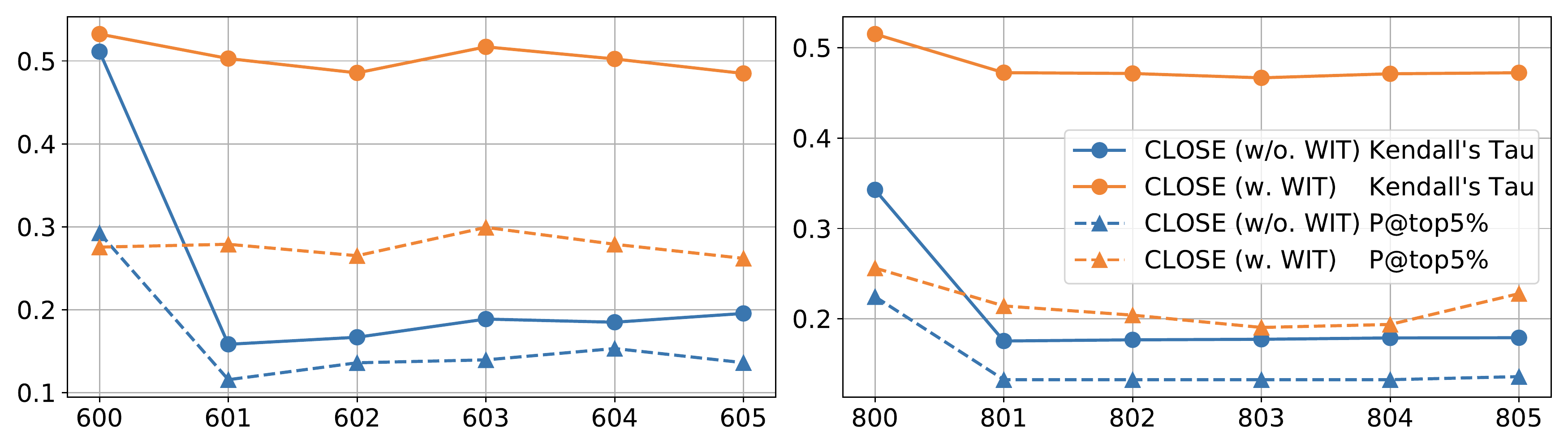}
    \caption{The trend of the ranking quality right after switching the sharing extent (at 600 and 800 epoch) on NAS-Bench-301. X-axis: Training epochs. Y-axis: Ranking quality (Kendall's Tau or P@top5\%). 
    }
    \label{fig:drop}
  \end{center}
\end{figure}

The results shown in Sec.~4.3 of the main paper demonstrate that WIT plays an important role in CLOSE. Here we provide a visualization to reveal its necessity more clearly. Fig.~\ref{fig:drop} shows that right after the curriculum (i.e., sharing extent) switch, randomly initializing the parameters of the new GLOW block and MLP unit significantly damages the ranking quality. On the contrary, WIT helps CLOSE to retain the high ranking quality.

\subsection{A Simplified Version of CLOSE: CLOSE-S}

In this section, we provide a simplified version of CLOSE (namely CLOSE-S). Based on the analysis in Sec.~3.1 of the main paper, we simply use the two sharing extents of Supernet-1 and Supernet-2 in two stages of the training process. Specifically, in the first stage (0 to 400 epoch), CLOSENet shares only one copy of parameters on all the edges in each cell-architecture. While in the second stage (400 to 1000 epoch), CLOSENet enables the operations on different edges to share different copies of parameters. The WIT and SRT are also adopted when switching the sharing scheme and extent at 400 epoch. In this way, we can easily apply CLOSE-S to train CLOSENet without the help of the GATE module, since the assignments of GLOW blocks are preset (but different) in these two stages. 
\begin{figure}[h!]
  \begin{center}
    \includegraphics[width=0.9\linewidth]{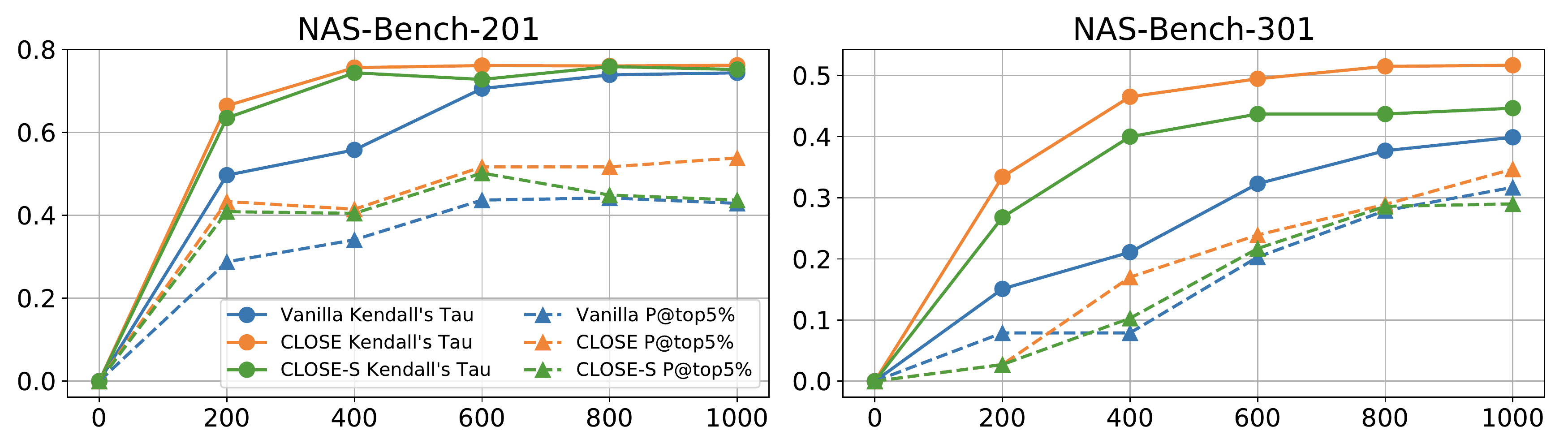}
    \caption{Comparison of the vanilla one-shot supernet, CLOSE and CLOSE-S. X-axis: Training epochs. Y-axis: Ranking quality (Kendall's Tau or P@top5\%).}
    \label{fig:simplify}
  \end{center}
\end{figure}

As shown in Fig.~\ref{fig:simplify}, CLOSE-S achieves a higher KD and P@top5\% than the vanilla supernet on two generic search spaces across the training process. Although it cannot reach the performances of CLOSE, CLOSE-S is easier to implement, thus can be adopted when the performance demand is not as strict.

\end{document}